\documentclass[table]{gtech}
\PassOptionsToPackage{table, usenames, dvipsnames}{xcolor}
\usepackage{xcolor}
\usepackage{amssymb}
\usepackage{multirow}
\usepackage{bigdelim}
\usepackage{longtable}
\usepackage{tabularray}
\usepackage{wrapfig}
\usepackage[most]{tcolorbox}
\usepackage{url}
\usepackage{float}
\usepackage{datatool}
\usepackage{enumitem}
\usepackage{subcaption} 
\usepackage[justification=centering]{caption}

\RequirePackage{tgpagella} 
\RequirePackage{mathpazo}  
\RequirePackage{inconsolata} 
\usepackage{makecell}
\usepackage{hyperref}
\usepackage{url}
\usepackage{times}
\usepackage{latexsym}
\usepackage[T1]{fontenc}
\usepackage[utf8]{inputenc}
\usepackage{microtype}
\usepackage{inconsolata}
\usepackage{amsmath}
\usepackage{amssymb}
\usepackage{xspace}
\usepackage{booktabs}
\usepackage{multirow}
\usepackage{wrapfig}
\usepackage{graphicx}
\usepackage{tabularx}
\usepackage{subcaption}
\usepackage{tcolorbox}
\usepackage{algorithm}
\usepackage{algpseudocode}
\usepackage{caption}
\usepackage{listings}

\newcommand{\paratitle}[1]{\vspace{1.5ex}\noindent\textbf{#1}}
\newcommand{\ie}{\emph{i.e.,}\xspace}

\newcommand{\eg}{\emph{e.g.,}\xspace}

\newcommand{\ignore}[1]{}
\newcommand{\OURS}{{\textbf{DEPO}}\xspace}

\newcommand{\OURSOff}{{\textbf{DEPO-Offline}}\xspace}

\algnewcommand{\Inputs}[1]{%
  \State \textbf{Inputs:}
  \Statex \hspace*{\algorithmicindent}\parbox[t]{0.98\linewidth}{\raggedright #1}
}
\algnewcommand{\Initialize}[1]{%
  \State \textbf{Initialize:}
  \Statex \hspace*{\algorithmicindent}\parbox[t]{0.98\linewidth}{\raggedright #1}
}

\renewcommand{\title}[1]{\newcommand{\titlelist}{{\huge\selectfont #1}}}

\title{\textbf{Towards High Data Efficiency in Reinforcement Learning with Verifiable Reward}}

\author[1,*]{Xinyu Tang}
\author[2,*]{Zhenduo Zhang}
\author[1]{Yurou Liu}
\author[1,\ddag]{Wayne Xin Zhao}
\author[2]{Zujie Wen}
\author[2]{Zhiqiang Zhang}
\author[2]{Jun Zhou}

\affiliation[1]{Gaoling School of Artificial Intelligence, Renmin University of China $^2$Ant Group \\[0.5em]}

\contribution[*]{Equal contribution.}
\contribution[\ddag]{Corresponding authors.}

\abstract{\fontsize{11pt}{12pt} \textit{
Recent advances in large reasoning models have leveraged reinforcement learning with verifiable rewards (RLVR) to improve reasoning capabilities.
However, scaling these methods typically requires extensive rollout computation and large datasets, leading to high training costs and low data efficiency.
To mitigate this issue, we propose \OURS, a \textbf{D}ata-\textbf{E}fficient \textbf{P}olicy \textbf{O}ptimization pipeline that combines optimized strategies for both offline and online data selection.
In the offline phase, we curate a high-quality subset of training samples based on diversity, influence, and appropriate difficulty.
During online RLVR training, we introduce a sample-level explorability metric to dynamically filter samples with low exploration potential, thereby reducing substantial rollout computational costs.
Furthermore, we incorporate a replay mechanism for under-explored samples to ensure adequate training, which enhances the model's final convergence performance.
Experiments across five reasoning benchmarks show that \OURS consistently outperforms existing methods in both offline and online data selection scenarios.
Notably, using only \textbf{20\%} of the training data, our approach achieves a \textbf{1.85} $\times$ speed-up on AIME24 and a \textbf{1.66} $\times$ speed-up on AIME25 compared to GRPO trained on the full dataset.
}}

\addsecondlogo[3.8cm]{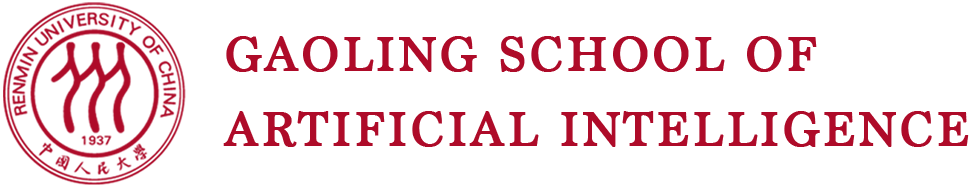} 
\begin{document}
\maketitle

\begin{figure}[ht]
    \centering
    \begin{subfigure}[b]{0.49\linewidth}
        \centering
        \includegraphics[width=\linewidth]{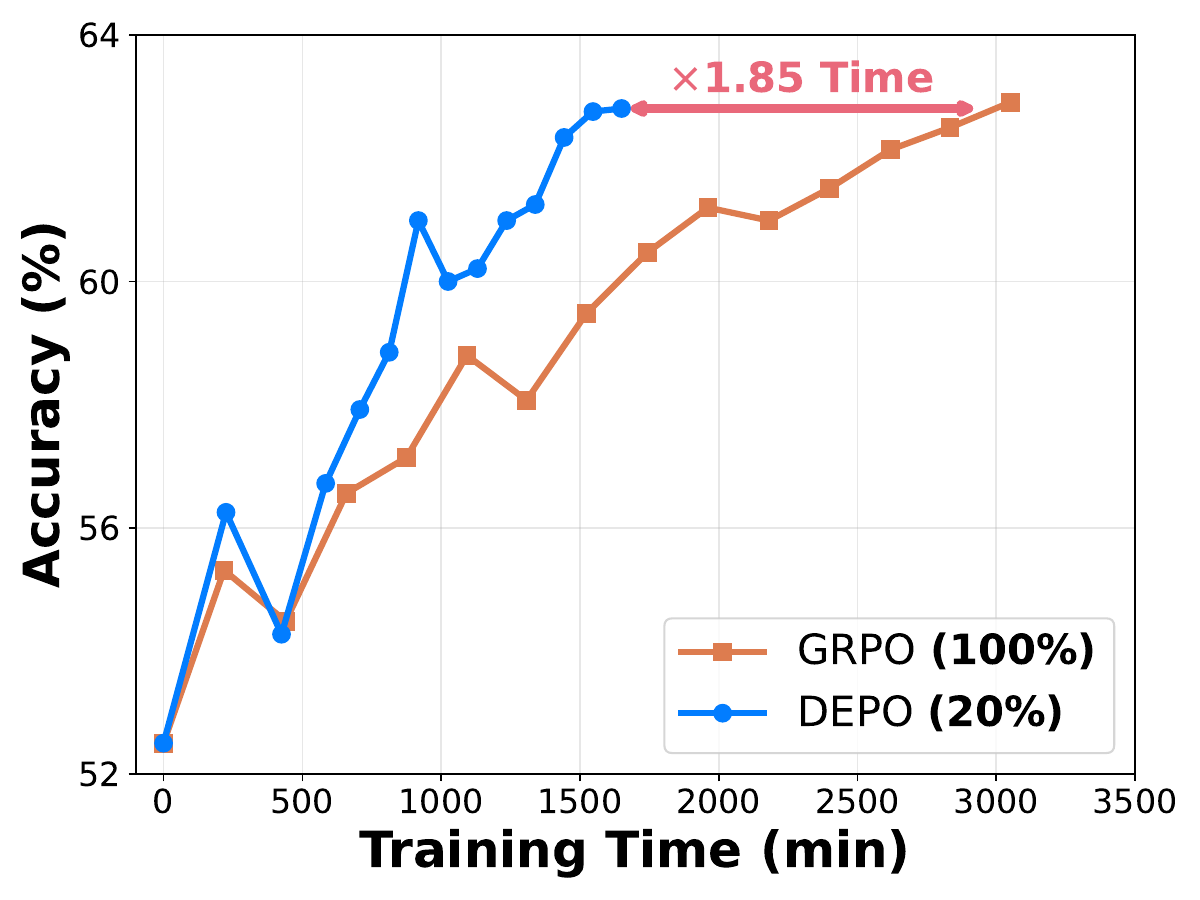}
        \caption{AIME24}
        \label{fig:accuracy_aime24}
    \end{subfigure}
    \begin{subfigure}[b]{0.49\linewidth}
        \centering
        \includegraphics[width=\linewidth]{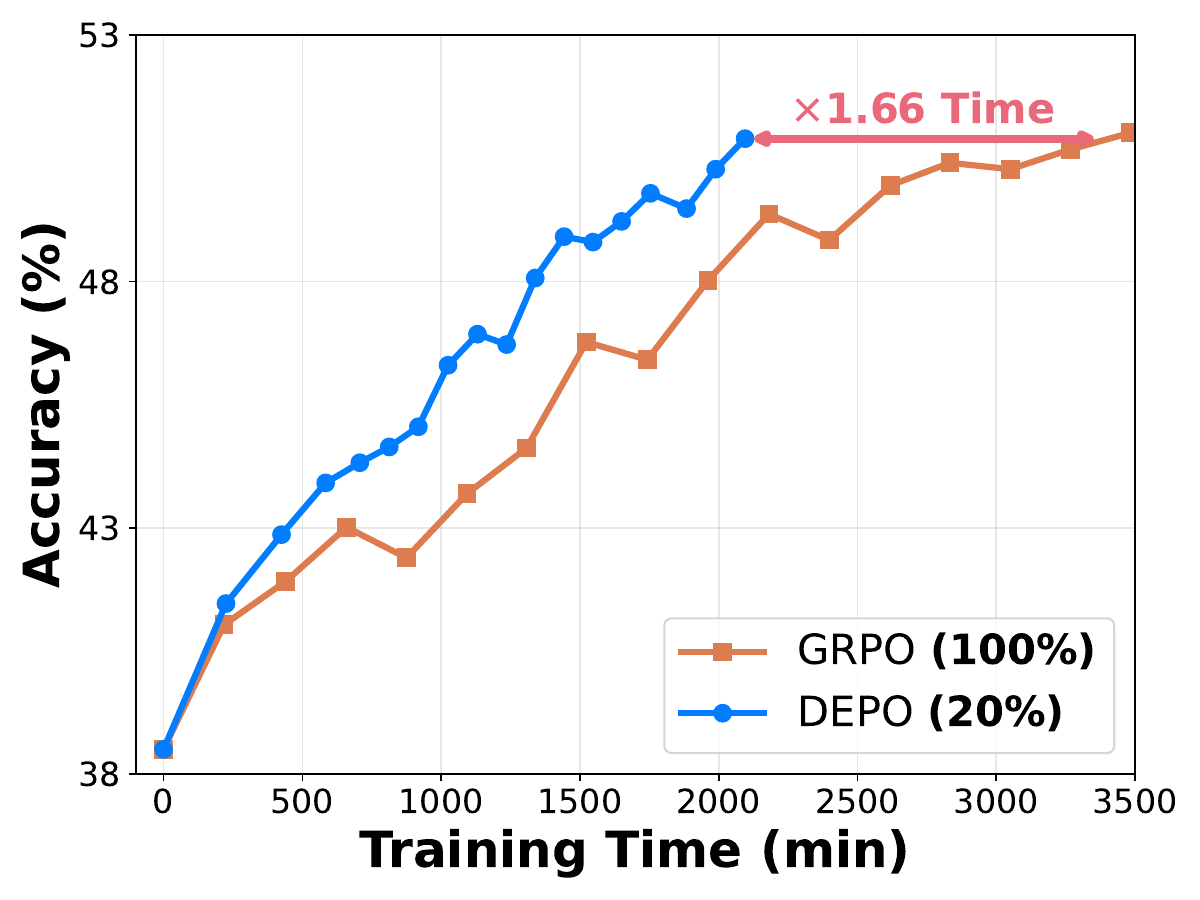}
        \caption{AIME25}
        \label{fig:accuracy_aime25}
    \end{subfigure}
    \caption{Accuracy of GRPO and \OURS during RLVR training. Our approach, using only 20\% of the training data, reduces training time by up to 1.6 times while achieving performance comparable to the GRPO baseline trained on the full dataset.}
\label{fig:intro}
\end{figure}

\section{Introduction}

Recent advancements in large reasoning models, such as OpenAI-o1~\citep{OpenAI-o1}, DeepSeek-R1~\citep{Deepseek-r1}, and Kimi~\citep{Kimi-1.5}, have drawn considerable attention due to their enhanced reasoning capabilities.
These models typically utilize reinforcement learning with verifiable rewards~(RLVR), including methods like PPO~\citep{PPO} and GRPO~\citep{GRPO}, to further unlock their reasoning potential.
In RLVR training, large language models~(LLMs) are required to explore multiple reasoning trajectories (\ie rollouts), and receive binary rewards based on the correctness of the final answer.
This process enables the models to iteratively refine their reasoning strategies.
A widely adopted approach to enhance RLVR performance is to scale the amount of training data and the number of rollouts.
This increased exploration allows LLMs to discover more diverse reasoning paths, leading to better convergence and improved performance.
Despite these advantages, this scaling strategy introduces significant drawbacks, that is, \textit{it substantially increases training cost and results in low data efficiency}.

To mitigate this problem, prior work has explored ways to improve data efficiency through both offline and online data selection strategies.
In the offline setting, ~\citet{LIMR} shows that even a small set of high-quality RLVR samples can effectively activate the reasoning abilities of LLMs.
Building on this, subsequent studies select RLVR data according to metrics such as historical reward trends~\citep{LIMR}, reward variance~\citep{1-shot-RLVR}, and gradient alignment~\citep{Learnalign}.
However, these approaches typically require prior training over the dataset to compute the metrics for selection, which entails high computational costs.
Other work~\citep{POLARIS} performs rollouts on the entire dataset and filters out samples that the model consistently answers correctly.
While this strategy can select beneficial samples, it still incurs significant computational costs due to conducting rollouts on the full dataset and does not account for interdependencies between training samples.
On the other hand, online data selection methods aim to improve the rollout efficiency, which is the major bottleneck in RL training.
\citet{GRESO} observes that a large portion of training data requires extensive rollout computations while contributing minimally to policy updates, which substantially reduces the training efficiency.
To mitigate this issue, it employs a probabilistic filter that excludes samples with zero historical reward variance.
Although this method reduces rollout costs effectively, it treats all historical non-zero variance samples equally and lacks a finer-grained metric to evaluate their exploration potential.

In this work, we propose a \textbf{D}ata-efficient \textbf{P}olicy \textbf{O}ptimization pipeline that integrates optimized strategies for both offline and online data selection, namely \OURS.
During the offline phase, we first apply a PageRank-weighted determinantal point process method to prune the dataset, preserving diverse and influential samples.
We then perform offline rollouts on this pruned subset and select samples whose difficulty scores approximate a normal distribution.
Consequently, we obtain a high-quality training subset that incorporates diversity, influence, and appropriate difficulty, which is subsequently used for RL training.
In the online RLVR training process, we propose a sample-level explorability metric to quantify the exploration potential of each sample based on its historical training dynamics.
Leveraging this metric, we strategically skip rollouts for samples with low explorability, thereby allocating computational resources for rollouts and policy updates on samples with higher exploration potential.
Moreover, to ensure all samples are adequately trained, we employ dynamic replay for under-explored samples to further improve the model's reasoning abilities.

To validate the effectiveness and efficiency of our approach, we conduct experiments on five reasoning benchmarks using three different LLMs.
Experimental results show that \OURS outperforms several competitive baselines in both offline and online data selection settings.
In particular, when using only \textbf{20\%} of the training data, \OURS achieves a \textbf{1.85} times speed-up on AIME24 and a \textbf{1.66} times speed-up on AIME25 with DeepSeek-R1-Distill-Qwen-7B compared to GRPO trained on the full dataset.



\section{Methodology}

\begin{figure*}[t]
    \centering
    \includegraphics[width=\textwidth]{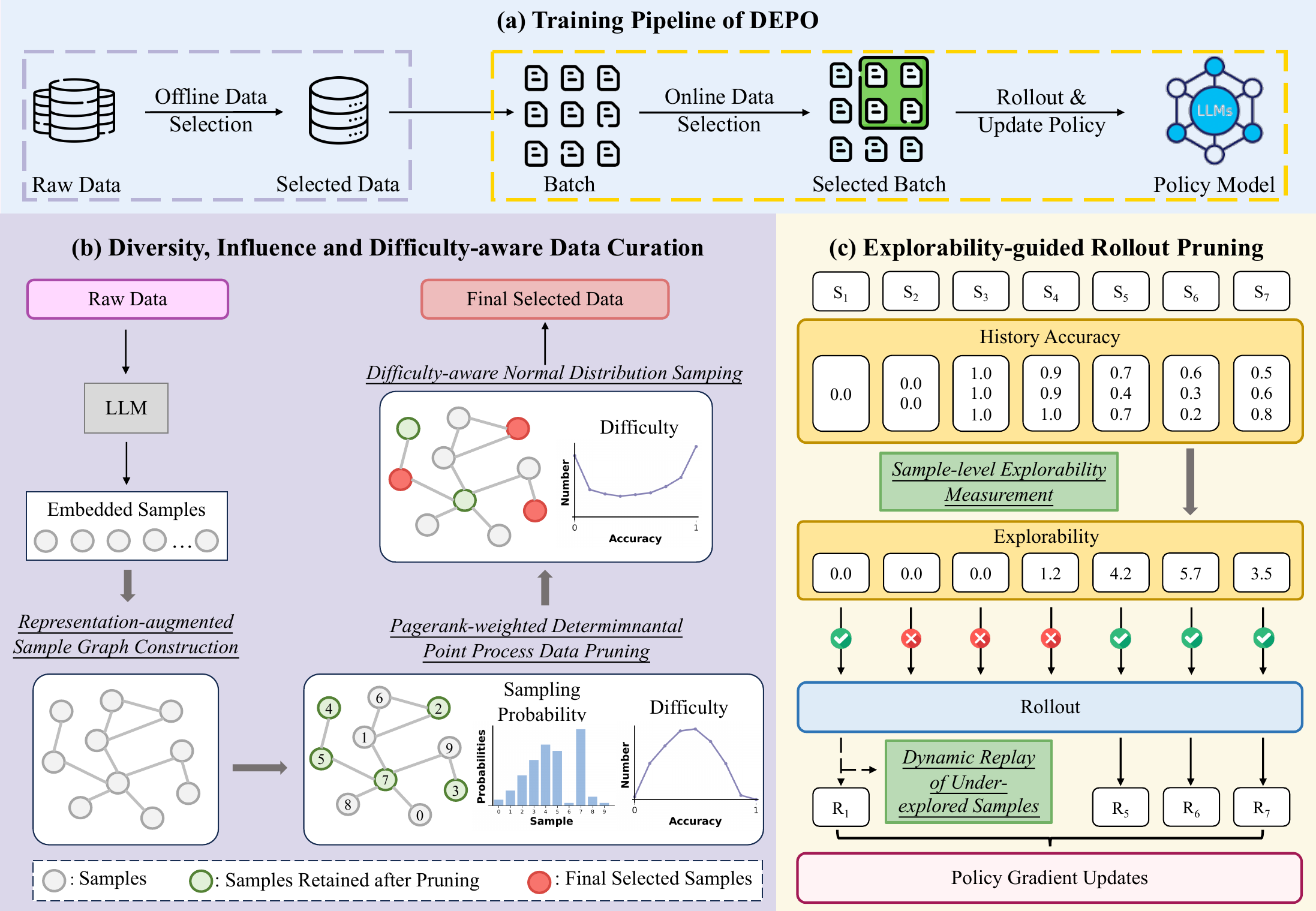}
    \caption{
        The overview of our proposed approach \OURS.
        Our approach improves data efficiency in the RLVR training pipeline (a) through a two-stage data selection strategy: (b) offline curation of a high-quality subset and (c) online rollout pruning guided by explorability.
        In the offline phase, we first construct a sample graph based on their representations, then apply PageRank-weighted Determinantal Point Process to select a diverse and influential subset, and finally sample from this subset with difficulty following a normal distribution.
        In the online phase, we assess the explorability of each sample based on its historical training dynamics and retain high-explorability ones for rollout, and actively replay under-explored samples to ensure sufficient training of all samples.
    }
\label{fig:main}
\end{figure*}

In this section, we present \OURS, an approach to enhance data efficiency in reinforcement learning with verifiable reward~(RLVR).
We first define the problem formulation, and then introduce a two-stage data selection process: (a) Diversity, Influence, and Difficulty-aware Data Curation for offline data selection, and (b) Explorability-guided Rollout Pruning for online rollout filtering.
Finally, we discuss the effectiveness and efficiency of our proposed approach.
The overall framework of \OURS is illustrated in Figure~\ref{fig:main}.

\subsection{Problem Formulation}

In this work, we study a large language model~(LLM) parameterized by $\theta \in \mathbb{R}^N$, which has been pretrained on large-scale corpora and will be further trained via reinforcement learning on an RLVR dataset $\mathcal{D} = \{x_i\}_{i=1}^{|\mathcal{D}|}$ to enhance the reasoning abilities.
Our primary goal is to optimize data efficiency by leveraging $\mathcal{D}$ more effectively without modification or augmentation, facilitating faster improvement in reasoning performance.
To achieve this, we first reduce data redundancy by selecting a high-quality subset offline.
Then, to further speed up RLVR training, we select samples with high exploration potential during training.
The following sections elaborate on these two stages.

\subsection{Diversity, Influence and Difficulty-aware Data Curation}

In this part, we present a data curation method to select a high-quality subset of RLVR data offline based on three criteria: diversity, influence, and difficulty.
First, we construct a sample graph based on their feature representations.
Next, we prune redundant samples by applying a PageRank-weighted Determinantal Point Process to retain a diverse and influential subset.
Finally, we further refine this subset by selecting samples whose difficulty levels follow a normal distribution.

\subsubsection{Representation-augmented Sample Graph Construction}

The first step of our approach is to model the relationships among samples using a graph structure.
Previous studies~\citep{task-vector,if-vector,long-cot-vector} demonstrate that internal model representations can effectively capture sample characteristics.
Inspired by this, we follow ~\citet{in-context-vector} and use the last token embedding from the final layer of the model as the sample representation, as it aggregates the model information and the entire input semantics.
Based on these representations, we construct a sample graph $\mathcal{G} = (\mathbf{V}, \mathbf{E}, \mathbf{P})$, where each vertex $v_i \in \mathbf{V}$ denotes a sample, an edge $e(i,j) \in \mathbf{E}$ connects node $v_i$ and $v_j$, and edge weight matrix $\mathbf{P}$ encodes the pairwise similarities between their embeddings.
This representation-augmented graph is subsequently used for our offline data selection process.

\subsubsection{Pagerank-weighted Determinantal Point Process Data Pruning}

We prune redundant data from the dataset by considering two sample properties: diversity and influence.
Diversity ensures broad information coverage and reduces redundancy, while influence reflects the importance of samples in the graph.
To promote diversity, we use Determinantal Point Process~(DPP)~\citep{DPP} to identify a subset that maximizes the determinant of the corresponding similarity submatrix: $\max_{Y \subseteq \mathbf{P}} \det(Y)$, where $Y$ is the submatrix of the full similarity matrix $\mathbf{P}$, and $\det()$ represents its determinant value.
Geometrically, this corresponds to selecting samples that form a larger volume in the feature space, with a larger volume indicating greater diversity.
To measure sample influence, we use PageRank~\citep{PageRank} to assign a weight $w_i$ to each sample, reflecting its representativeness.
Our goal is to maximize the influence of the selected samples: $\max_{Y \subseteq \mathbf{P}} \prod\limits_{i \in Y} w_i$.
To unify both objectives, we combine diversity and influence into a single kernel matrix $L_Y$ and maximize its determinant:
\begin{align}
\max_{Y} \Big(\det(Y) \cdot \prod_{i \in Y} w_i \Big) 
&= \max_{Y} \det\left( 
\underbrace{\operatorname{diag}(\mathbf{w}_Y^{1/2}) \cdot Y \cdot \operatorname{diag}(\mathbf{w}_Y^{1/2})}_{L_Y}
\right),
\label{eq:influence_diversity}
\end{align}
where $L_Y$ denotes the kernel matrix that integrates both diversity and influence of the selected subset.
We provide the detailed theoretical analysis of this optimization objective in Appendix~\ref{sec:theoretic_proof}.

Since the optimization problem is NP-hard, we follow \citet{DPP1,DPP2} and employ an efficient greedy algorithm to approximate the solution.
The algorithm iteratively selects samples based on a dynamically updated probability distribution and updates the features of the remaining samples via Gram-Schmidt orthogonalization.
The pseudo-code of this algorithm is presented in Algorithm~\ref{alg:dpp-sampling}.

\subsubsection{Difficulty-aware Normal Distribution Sampling}

After pruning data via PageRank-weighted DPP, we obtain a representative and diverse subset $Y$.
However, in RLVR training, samples that are too easy or too hard provide limited learning signals and contribute little to policy optimization.
To better align the training data with the current model’s capability, we propose a difficulty-aware sampling strategy that prioritizes samples of moderate difficulty.
Specifically, for each sample $i$ in subset $Y$, we generate $G$ offline trajectories using the current LLM policy $\pi_{\theta}$ and compute its accuracy $\text{Acc}_i$:
\begin{align}
   \text{Acc}_i = \mathbb{E}_{\{o_j\}_{j=1}^G \sim \pi_{\theta}} \left[ \frac{1}{G} \sum_{j=1}^G \mathcal{V}(o_j, a_i) \right].
\end{align}
Here, $\{o_j\}_{j=1}^G \sim \pi_{\theta}$ are $G$ responses generated by $\pi_{\theta}$ for question $i$, and $\mathcal{V}(o_j, a_i)$ denotes a verifier that evaluates whether model output $o_j$ matches the ground-truth answer $a_i$.

Using the accuracy scores as a measure of difficulty, we then sample from $Y$ according to a normal distribution $\mathcal{N}(\mu, \sigma^2)$, where $\mu$ and $\sigma$ are the mean and standard deviation of the accuracies in the final selected subset.
Thus, the sampling probability for each sample is proportional to the standard normal density function, which assigns higher probabilities to samples near the mean difficulty:
\begin{align}
    p_i = \frac{ \phi\left( \frac{\text{Acc}_i - \mu}{\sigma} \right) }{ \sum_{k \in Y} \phi\left( \frac{\text{Acc}_k - \mu}{\sigma} \right) }, \quad \text{where} \quad \phi(x) = \frac{1}{\sqrt{2\pi}} e^{-x^2/2}.
\end{align}
The final RL training subset $D^{\text{sub}}$ is constructed by sampling from $Y$ according to the probability distribution $\mathcal{P} = \{p_i\}_{i \in Y}$, which is subsequently used for our RLVR training.

\subsection{Explorability-guided Rollout Pruning}

In RLVR training, it is notable that rollout generation is computationally expensive and becomes the main bottleneck for training speed.
To further improve efficiency, we propose an explorability metric to quantify the exploration potential of samples.
This metric prioritizes high explorability samples for rollout generation and policy gradient updates, thereby reducing computational cost and enhancing data efficiency by selectively skipping unnecessary rollouts.
Additionally, to avoid missing under-explored samples that may become valuable, we implement a dynamic replay mechanism to revisit the least-explored samples in the training process.
The pseudo-code of this method is presented in Algorithm~\ref{alg:explorability-pruning}.

\subsubsection{Sample-level Explorability Measurement}

In RLVR training, high-entropy samples promote exploration, while training on low-entropy samples may lead to overfitting and reward hacking.
Therefore, we propose an explorability metric to identify samples that have high entropy in recent epochs, as they are more likely to yield diverse reasoning paths and improve model performance efficiently.
To mitigate instability from extremely high-entropy negative samples (\ie, pathological trajectories), we apply a threshold $\lambda$ to exclude them.
Formally, the explorability metric $\mathcal{E}$ aggregates entropy-based signals over a sliding window of recent epochs, combining both positive samples and negative samples with entropy below the threshold:
\begin{equation}
\begin{aligned}
\mathcal{E}(q,a,\{O^t\}_{t=e-w+1}^e) &= \frac{1}{w} \sum_{t=e-w+1}^{e} \frac{1}{G} \sum_{i=1}^G E(q,a,o^t_i), \\ 
E(q,a,o^t_i) =  \hat{A}_{i} \cdot E(o_i^t) \cdot \mathbb{I} & \left[ \mathcal{V}(o_i^t, a) = 1 \lor \left(\mathcal{V}(o_i^t,a)=0 \land E(o_i^t) \leq \lambda E(\overline{o_i^{t+}}) \right) \right] , 
\label{eq:explorability}
\end{aligned}
\end{equation}
where $q$ and $a$ denote the question and the answer of the sample, $O^t  = \{o_i^t\}_{i=1}^G $ is the rollouts in epoch $t$, $w$ is the window size of recent epochs, $E(o^t_i)$ denotes the average entropy of rollout $o^t_i$ across all tokens, $E(\overline{o_i^+})$ is the mean entropy of all positive samples, and $\hat{A}_{i} = \frac{r_i - \text{mean}(\{r_i\}_{i=1}^G)}{\text{std}(\{r_i\}_{i=1}^G)}$ is the normalized reward across the group.

\subsubsection{Dynamic Replay of Under-explored Samples}

During training, samples with consistently low explorability may be overlooked, even if they become valuable in later stages.
To mitigate this issue, we introduce a dynamic replay mechanism that intentionally incorporates under-explored samples.
Specifically, each training batch consists of two types of samples:
(1) samples ranked in the top-$\alpha_e\%$ by explorability, and
(2) the $\rho\%$ of samples that have been least explored throughout the training.
Thus, the optimization objective for \OURS is:
\begin{equation}
\begin{aligned}
\mathcal{J}_\text{\OURS}(\theta)& = \mathbb{E}_{\textcolor{red}{\mathcal{B} \sim \mathcal{D}, (q,a)\sim \mathcal{B}}, \{o_i\}_{i=1}^G\sim \pi_{\theta_\text{old}}(\cdot\mid q)}
\Bigg[ \textcolor{red}{\mathbb{I} \Big[ q,a,\{O^t\}_{t=e-w+1}^e \Big]} \frac{1}{G}\sum_{i=1}^{G} \cdot \frac{1}{|o_i|}\sum_{t=1}^{|o_i|} \\&
  \Bigg( 
\min \Big( r_{i,t}(\theta) \hat{A}_{i},  
\ \text{clip} \Big( r_{i,t}(\theta), 1 - \varepsilon, 1 + \varepsilon \Big) \hat{A}_{i} \Big)
- \beta D_{\text{KL}}(\pi_{\theta} || \pi_{\text{ref}}) 
\Bigg) \Bigg],
\label{eq:oursloss}
\end{aligned}
\end{equation}
where
\begin{align*}
\mathbb{I} \Big[ q,a,\{O^t\}_{t=e-w+1}^e \Big] = 
\begin{cases}
1 & \mathcal{E}(q, a, \{O^t\}_{t=e-w+1}^e) \ \text{is} \ \text{top-}\alpha_e\%  \\
1 & |\{O^t\}_{t=1}^e| \ \text{is} \ \text{bottom-}\rho\% \\
0 & \text{else},
\end{cases}
\end{align*}
\begin{equation}
    r_{i,t}(\theta)=\frac{\pi_{\theta}(o_{i,t} \mid q, o_{i,<t})}{\pi_{\theta_{\text{old}}}(o_{i,t} \mid q,o_{i,<t})}, \quad\text{and} \quad \alpha_e = \alpha_0 - d \cdot e.
\end{equation}
To adaptively shift the training focus from broad exploration in the early stages to specialized refinement later, we progressively reduce the proportion of high-explorability samples per epoch using a linear decay rate $d$.
Here, $\alpha_0$ denotes the initial sampling rate, $d$ is the decay rate, and $e$ is the current epoch.
The online rollout pruning strategy optimizes computational resource allocation by focusing on samples with high exploration potential, and strategically replaying under-explored samples to ensure all samples are sufficiently trained, significantly improving training efficiency while maintaining comparable model performance.

\subsection{Discussion}


\paratitle{Effectiveness of \OURS.}
Existing offline RLVR data selection methods typically curate datasets prior to RL training.
Several representative approaches select samples based on metrics such as reward trends~\citep{LIMR}, reward variance~\citep{1-shot-RLVR}, and gradient alignment~\citep{Learnalign}.
However, these methods rely solely on early-stage training dynamics, which limits their effectiveness in later training phases.
Moreover, such methods primarily focus on the distribution of data difficulty while neglecting interrelationships among samples.
In contrast, \OURS selects RLVR subsets offline by jointly optimizing three key criteria: diversity, influence, and difficulty.
On the other hand, online RLVR data selection methods dynamically filter data during training.
For instance, GRESO~\citep{GRESO} probabilistically removes samples with historical zero-variance rewards.
While effective, this approach treats all historical non-zero variance samples equally and may underperform in scenarios where zero variance samples are scarce.
Our method overcomes these limitations by evaluating samples based on their explorability, actively prioritizing those with higher exploration potential for rollouts and policy gradient updates.

\paratitle{Efficiency of \OURS.}
Compared to existing offline methods that require training on the full or warmup dataset for multiple epochs to guide data selection~\citep{LIMR,1-shot-RLVR, Learnalign}, or performing offline rollouts across the entire dataset to evaluate sample difficulty~\citep{POLARIS}, \OURS adopts a more computationally efficient selection strategy.
Our approach first prunes the dataset based on diversity and influence, and then performs offline rollouts on this reduced subset to evaluate difficulty.
This two-stage selection process eliminates the need for additional RLVR training and significantly reduces the rollout costs.
Online methods typically incur high computational overhead, as they require generating extensive rollouts before policy gradient updates~\citep{DAPO}.
In contrast, \OURS alleviates this burden by estimating sample explorability directly from historical training dynamics, thereby minimizing rollout costs and enhancing training efficiency.
\begin{table*}[t]
\centering
\Large
\caption{Performance comparison of various methods using Deepseek-R1-Distill-Qwen-7B, Deepseek-R1-Distill-Llama-8B, and Qwen2.5-7B-Math. ``\textbf{Offline}'' and ``\textbf{Online}'' refer to the offline and online data selection methods, respectively. ``Ratio, ``Time'', and ``RN'' denote the ratio of selected data, total training time, and total rollout numbers, respectively.}
\resizebox{\textwidth}{!}{
\begin{tabular}{c|cr|cccccc|ccc}
    \toprule
    \multirow{2.5}{*}{\begin{tabular}[c]{@{}c@{}}\textbf{Model}\end{tabular}} & \multicolumn{2}{c|}{\multirow{2.5}{*}{\begin{tabular}[c]{@{}c@{}}\textbf{Method}\end{tabular}}} & \multicolumn{6}{c|}{\textbf{Accuracy}} & \multicolumn{3}{c}{\textbf{Efficiency}} \\
    \cmidrule{4-12}
    & & & \textbf{AIME 24} & \textbf{AIME 25} & \textbf{MATH500} & \textbf{GPQA} & \textbf{LiveCodeBench}  & \textbf{Average} & \textbf{Ratio} & \textbf{Time} & \textbf{RN} \\
    \midrule
    \multirow{11}{*}{\begin{tabular}[c]{@{}c@{}}\textbf{Deepseek-R1-}\\\textbf{Distill-Qwen-7B}\end{tabular}} & 
    \multirow{2}{*}{\begin{tabular}[c]{@{}c@{}}-\end{tabular}} 
      & Base              & 51.5 & 38.5 & 91.2 & 45.9 & 37.0 & 52.8 & - & - & - \\
    & & Full              & 63.4 & 52.7 & 96.3 & 51.6 & 44.7 & 61.7 &  100\% &  100\% &  100\% \\
    \cmidrule{2-12}
    & \multirow{6}{*}{\begin{tabular}[c]{@{}c@{}}\textbf{Offline}\end{tabular}} 
      & Random            & 56.8 & 43.0 & 94.6 & 47.3 & 39.6 & 56.3 &  20\% &  98\% &  100\% \\
    & & PPL-Top           & 57.5 & 45.3 & 95.0 & 48.2 & 40.4 & 57.3 &  20\% &  101\% &  100\% \\
    & & PPL-Middle        & 57.5 & 45.4 & 95.2 & 48.6 & 40.8 & 57.5 &  20\% &  97\% &  100\% \\
    & & LIMR              & 59.9 & 46.4 & 95.1 & 48.5 & 40.9 & 58.2 &  20\% &  99\% &  100\% \\
    & & Learnalign        & 60.1 & 46.8 & 95.5 & 49.0 & 41.9 & 58.7 &  20\% &  102\% &  100\% \\
    & & \OURSOff          & \textbf{63.1} & \textbf{51.7} & \textbf{96.1} & \textbf{51.7} & \textbf{44.5} &  \textbf{61.4} &  20\% &  99\% &  100\% \\
    \cmidrule{2-12}
    & \multirow{3}{*}{\begin{tabular}[c]{@{}c@{}}\textbf{w Online}\end{tabular}} 
      & + Random          & 58.7 & 45.3 & 93.1 & 47.2 & 39.3 & 56.7 &  20\% &  58\% &  40\% \\
    & & + GRESO           & 60.2 & 47.4 & 94.3 & 48.1 & 40.6 & 58.1 &  20\% &  55\% &  40\% \\
    & & + \OURS           & \textbf{62.8} & \textbf{50.9} & \textbf{95.9} & \textbf{51.4} & \textbf{44.3} &  \textbf{61.1} &  20\% &  57\% &  40\% \\
    \midrule

    \multirow{11}{*}{\begin{tabular}[c]{@{}c@{}}\textbf{Deepseek-R1-}\\\textbf{Distill-Llama-8B}\end{tabular}} & 
    \multirow{2}{*}{\begin{tabular}[c]{@{}c@{}}-\end{tabular}} 
      & Base              & 41.1 & 30.4 & 88.5 & 37.3 & 44.3 & 48.3 & - & - & - \\
    & & Full              & 56.9 & 45.1 & 94.8 & 44.4 & 49.6 & 58.2 &  100\% &  100\% &  100\% \\
    \cmidrule{2-12}
    & \multirow{6}{*}{\begin{tabular}[c]{@{}c@{}}\textbf{Offline}\end{tabular}} 
      & Random            & 47.6 & 38.7 & 90.6 & 39.6 & 44.5 & 52.2 &  20\% &  100\% &  100\% \\
    & & PPL-Top           & 48.2 & 39.3 & 90.4 & 39.1 & 45.0 & 52.4 &  20\% &  102\% &  100\% \\
    & & PPL-Middle        & 49.9 & 39.2 & 91.1 & 39.4 & 45.7 & 53.1 &  20\% &  102\% &  100\% \\
    & & LIMR              & 52.3 & 40.9 & 91.6 & 41.0 & 45.3 & 54.2 &  20\% &  97\% &  100\% \\
    & & Learnalign        & 54.7 & 41.8 & 91.6 & 40.8 & 46.2 & 55.0 &  20\% &  98\% &  100\% \\
    & & \OURSOff          & \textbf{57.6} & \textbf{44.8} & \textbf{94.2} & \textbf{43.6} & \textbf{49.3} & \textbf{57.9} &  20\% &  100\% &  100\% \\
    \cmidrule{2-12}
    & \multirow{3}{*}{\begin{tabular}[c]{@{}c@{}}\textbf{w Online}\end{tabular}} 
      & + Random          & 50.2 & 38.4 & 90.1 & 39.7 & 44.8 & 52.6 &  20\% &  55\% &  40\% \\
    & & + GRESO           & 52.6 & 40.2 & 92.0 & 40.5 & 46.6 & 54.4 &  20\% &  54\% &  40\% \\
    & & + \OURS           & \textbf{56.8} & \textbf{44.4} & \textbf{93.7} & \textbf{42.8} & \textbf{48.8} & \textbf{57.3} &  20\% &  56\% &  40\% \\
    
    \midrule

    \multirow{11}{*}{\begin{tabular}[c]{@{}c@{}}\textbf{Qwen2.5-}\\\textbf{7B-Math}\end{tabular}} & 
    \multirow{2}{*}{\begin{tabular}[c]{@{}c@{}}-\end{tabular}} 
      & Base              & 13.4 & 6.4  & 54.5 & 28.7 & 5.6 & 21.7 & - & - & - \\
    & & Full              & 30.2 & 20.3 & 86.8 & 35.7 & 13.6 & 37.3 &  100\% &  100\% &  100\% \\
    \cmidrule{2-12}
    & \multirow{6}{*}{\begin{tabular}[c]{@{}c@{}}\textbf{Offline}\end{tabular}} 
      & Random            & 22.5 & 13.3 & 72.5 & 30.3 & 8.2 & 29.4 &  20\% &  98\% &  100\% \\
    & & PPL-Top           & 24.1 & 13.8 & 76.2 & 31.0 & 9.6 & 30.9 &  20\% &  102\% &  100\% \\
    & & PPL-Middle        & 24.8 & 14.3 & 76.0 & 30.6 & 9.9 & 31.1 &  20\% &  98\% &  100\% \\
    & & LIMR              & 26.5 & 15.8 & 78.0 & 32.2 & 10.6 & 32.6 &  20\% &  101\% &  100\% \\
    & & Learnalign        & 27.1 & 17.2 & 80.5 & 33.6 & 10.9 & 33.9 &  20\% &  98\% &  100\% \\
    & & \OURSOff          & \textbf{30.0} & \textbf{19.4} & \textbf{85.8} & \textbf{35.2} & \textbf{13.2} & \textbf{36.7} &  20\% &  99\% &  100\% \\
    \cmidrule{2-12}
    & \multirow{3}{*}{\begin{tabular}[c]{@{}c@{}}\textbf{w Online}\end{tabular}} 
      & + Random          & 24.3 & 14.5 & 74.3 & 31.1 & 9.3 & 30.7 &  20\% &  57\% &  40\% \\
    & & + GRESO           & 27.7 & 16.8 & 80.7 & 33.4 & 10.9 & 33.9 &  20\% &  57\% &  40\% \\
    & & + \OURS           & \textbf{29.8} & \textbf{19.2} & \textbf{86.3} & \textbf{34.8} & \textbf{12.8} & \textbf{36.6} &  20\% &  55\% &  40\% \\
    
    \bottomrule
\end{tabular}
}
\label{tab:main_results}
\end{table*}

\section{Experiments}

In this section, we first describe the experimental setup, then present the main results and provide a detailed analysis.

\subsection{Experimental Setup}

We run our experiments on DeepSeek-R1-Distill-Qwen-7B~\citep{Deepseek-r1}, DeepSeek-R1-Distill-Llama-8B~\citep{Deepseek-r1}, Qwen2.5-Math-7B~\citep{Qwen2.5-Math-7B}.
Our method is implemented based on the Verl~\citep{verl} pipeline and uses vLLM~\citep{vllm} for rollout.
For training, we use DAPO-Math~\citep{DAPO} as the training dataset and apply the GRPO algorithm to train the models.
For evaluation benchmarks, we use three widely used mathematical reasoning benchmarks (\ie AIME24, AIME25, Math500~\citep{Math500}) and two other reasoning benchmarks (\ie GPQA~\citep{GPQA} and LiveCodeBench~\citep{LiveCodeBench}) to evaluate the model performance.
We follow ~\citet{GRESO} to evaluate models on these benchmarks every 50 steps and report the performance of the checkpoint that obtains the best average performance on five benchmarks.
More experimental details are provided in Appendix~\ref{app:detailed_exp_setup}.
To facilitate a systematic comparison, we select several representative offline and online data selection methods as baselines.
For offline data selection methods, we compare our method with random selection, conventional supervised fine-tuning~(SFT) data selection methods (\ie PPL-Top~\citep{PPL-Top} and PPL-Middle~\citep{PPL-Middle}), and the RLVR selection methods (\ie LIMR~\citep{LIMR} and Learnalign~\citep{Learnalign}).
For online data selection methods, we integrate them into our offline selected subset and compare our method with random online selection and GRESO~\citep{GRESO}.
Detailed descriptions of these baselines are provided in Appendix~\ref{app:baselines}.

\subsection{Main Results}

\begin{figure}[t]
    \centering
    \begin{subfigure}[b]{0.24\linewidth}
        \centering
        \includegraphics[width=\linewidth]{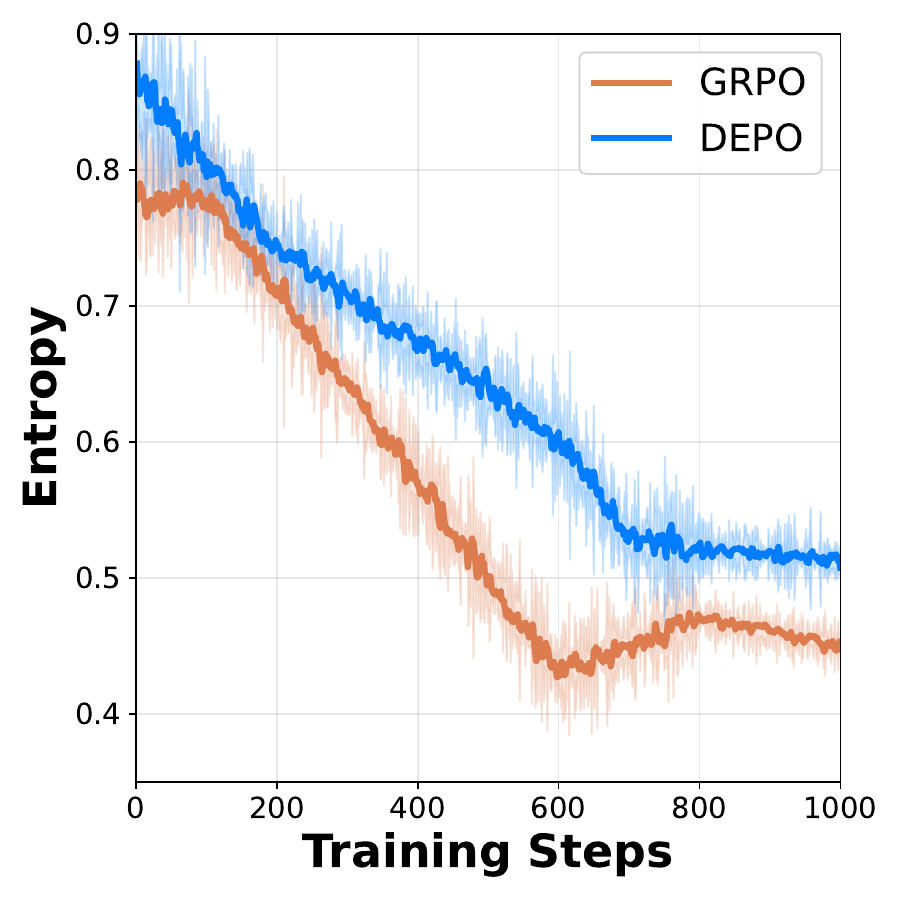}
        \caption{Entropy}
        \label{fig:entropy}
    \end{subfigure}
    \begin{subfigure}[b]{0.24\linewidth}
        \centering
        \includegraphics[width=\linewidth]{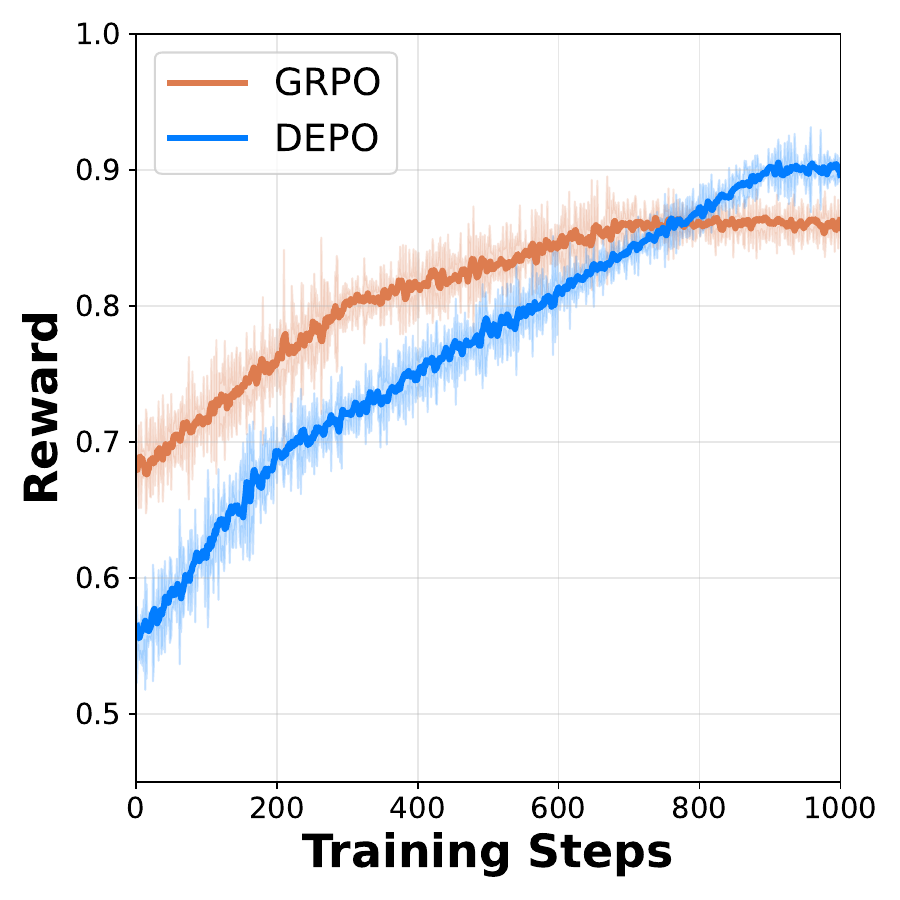}
        \caption{Reward}
        \label{fig:reward}
    \end{subfigure}
    \begin{subfigure}[b]{0.24\linewidth}
        \centering
        \includegraphics[width=\linewidth]{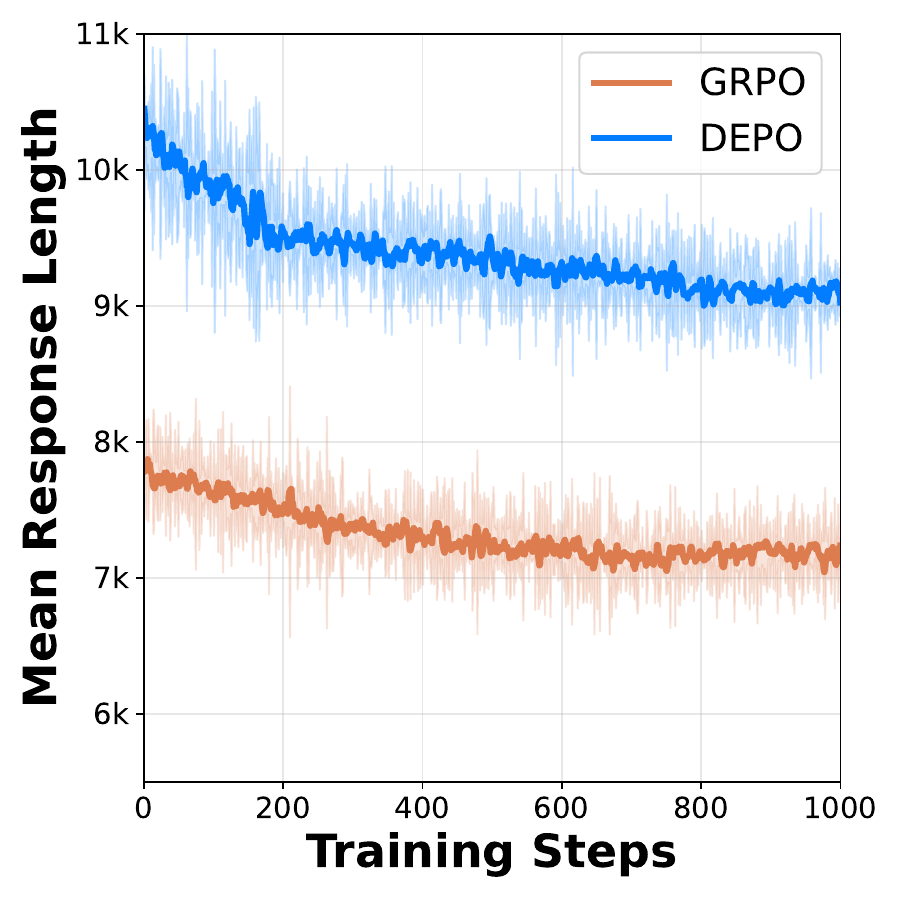}
        \caption{Response Length}
        \label{fig:response_length}
    \end{subfigure}
    \begin{subfigure}[b]{0.24\linewidth}
        \centering
        \includegraphics[width=\linewidth]{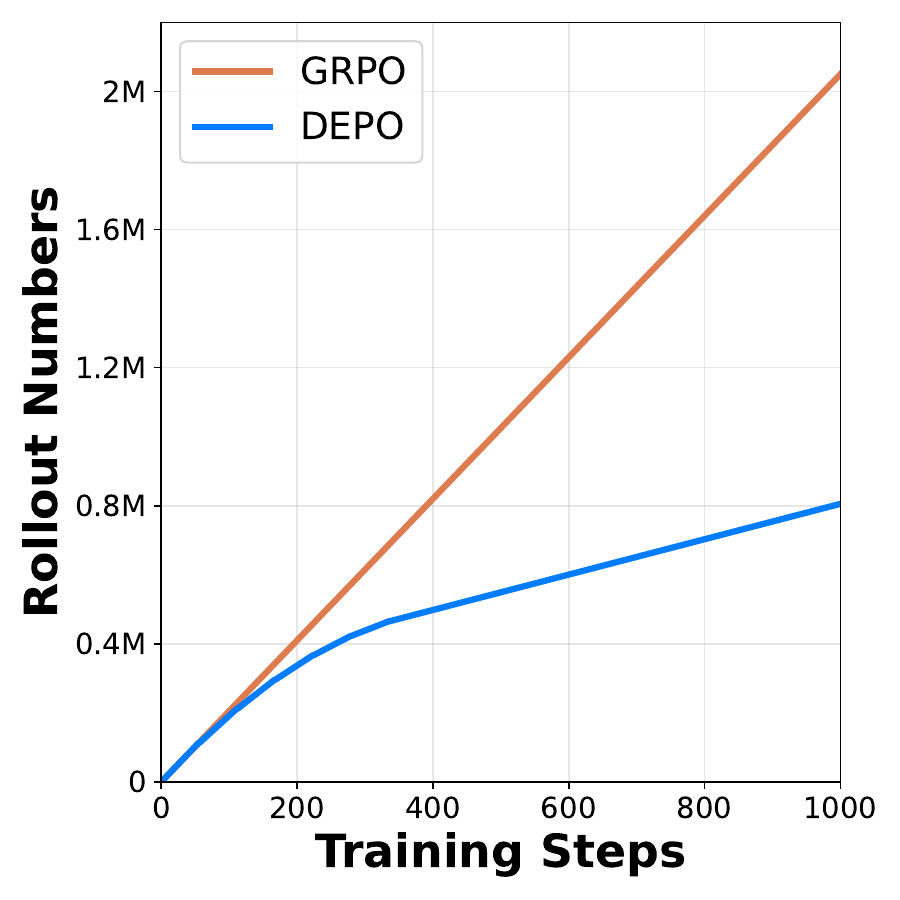}
        \caption{Rollout Numbers}
        \label{fig:total_sampling_num}
    \end{subfigure}
    \caption{The RLVR training dynamics of GRPO and \OURS.}
\label{fig:training_dynamic}
\end{figure}

In this section, we present the main results of our experiments.
The training dynamics of \OURS and GRPO are illustrated in Figure~\ref{fig:training_dynamic}.
The main results are presented in Table~\ref{tab:main_results}.

\paratitle{\OURS selects high-quality subsets offline for RLVR training.}
As shown in Table~\ref{tab:main_results}, reinforcement learning on mathematical data not only enhances the model’s mathematical reasoning capabilities but also improves its performance on other reasoning tasks (\ie GPQA and LiveCodeBench).
Regarding offline data selection methods, although supervised fine-tuning data selection methods (\ie PPL-Top and PPL-Middle) incorporate additional information from training data, their performance remains relatively poor.
One possible reason is the mismatch between SFT and RL objectives.
SFT aims to maximize the likelihood of target outputs, making perplexity a natural indicator of sample difficulty.
In contrast, RL focuses on reward maximization, which requires the difficulty of training samples to match the model's current capability.
Moreover, RLVR methods (\ie LIMR and Learnalign), which perform training on RLVR data prior to selection, lead to further improvements.
However, these approaches tend to select samples that match the initial model capabilities, resulting in improvements during early stages but limiting performance in later phases.
Additionally, such methods often overlook interdependencies among problems.
In comparison, \OURS achieves the best performance across all methods, nearly matching the performance of training on the full dataset.
One possible reason is that \OURS selects samples with diversity, influence, and a normally distributed range of difficulty, ensuring rapid improvement during the early stage.
Besides, the diversity in data difficulty supports sustained progress in later stages.
As illustrated in Figure~\ref{fig:training_dynamic}, our method selects samples with higher initial entropy, lower initial rewards, and longer response lengths.
This ensures that our selected data matches the model’s current capability and offers diverse exploration paths, thereby facilitating more effective RLVR training.

\paratitle{\OURS saves training computational costs and maintains comparable performance.}
As shown in Table~\ref{tab:main_results}, randomly reducing the number of rollouts during training leads to a significant performance degradation, indicating that the selection of samples for rollout has a substantial impact on the results.
Moreover, GRESO improves model performance by probabilistically filtering out historical zero-variance samples, which contribute little to training.
However, this filtering strategy does not account for the differences among historical non-zero-variance samples, which limits its overall performance.
In comparison, \OURS achieves performance comparable to full rollouts while using less than 60\% of the training time and 40\% of the rollout budgets.
This is because our method dynamically estimates the explorability of samples based on historical training dynamics, prioritizing those with higher explorability for rollout and policy updates.
Furthermore, we incorporate a replay strategy for under-explored samples to ensure sufficient training across all data.
As illustrated in Figure~\ref{fig:training_dynamic}, our rollout pruning strategy consistently selects samples with higher entropy, faster reward growth, and longer generated responses throughout the training process, demonstrating that \OURS efficiently improves reasoning performance.

\subsection{Ablation Studies}

\begin{wraptable}{r}{0.5\textwidth}
\centering
\Large
\caption{Ablation study on three math benchmarks.}
\resizebox{0.5\textwidth}{!}{

\begin{tabular}{l | c c c}
\toprule
\cmidrule{2-4}
Dataset & AIME 24 & AIME 25 & Math500 \\
\midrule
\OURS & \textbf{62.8} & \textbf{50.9} & \textbf{95.9} \\
\midrule
\midrule
\multicolumn{4}{c}{\textit{Offline Data Selection}} \\
\midrule
\midrule
w/o Pagerank-weighted DPP & 62.1 & 50.0 & 95.6 \\
w/o Difficulty-aware Sampling  & 60.3 & 47.8 & 95.1 \\
\midrule
\midrule
\multicolumn{4}{c}{\textit{Online Data Selection}} \\
\midrule
\midrule
w/o Explorability Measurement & 58.7 & 45.3 & 93.1 \\
w/o Under-explored Sample Replay & 62.3 & 48.4 & 95.2 \\
\bottomrule
\end{tabular}
}
\vspace{-0.2cm} 
\label{tab:ablation}
\end{wraptable}

To evaluate the effectiveness of each component in our method, we conduct ablation studies on three math benchmarks using DeepSeek-R1-Distill-Qwen-7B.
As shown in Table~\ref{tab:ablation}, removing any component can lead to performance degradation, demonstrating that all components are essential to our approach.
In the offline data selection phase, removing Difficulty-aware Sampling results in the most significant performance drop, which indicates that sample difficulty is a crucial factor for selection.
For online selection, replacing explorability-based filtering with random filtering causes a substantial decline in performance.
This confirms that explorability is an important indicator of a sample’s potential value in RLVR training.
Furthermore, we find that removing Under-explored Sample Replay notably impairs performance on more challenging tasks such as AIME 25 (performance drops from 50.9 to 48.4), suggesting that replaying challenging and under-trained samples is beneficial for enhancing the model’s ability to solve harder problems.

\subsection{Detailed Analysis}

In this part, we conduct a detailed analysis of our proposed method.
Unless otherwise specified, we report the average accuracy across five benchmarks using DeepSeek-R1-Distill-Qwen-7B.

\begin{figure}[t]
    \centering
    \begin{minipage}[b]{0.49\linewidth}
        \centering
        \begin{subfigure}[b]{0.49\linewidth}
            \centering
            \includegraphics[width=\linewidth]{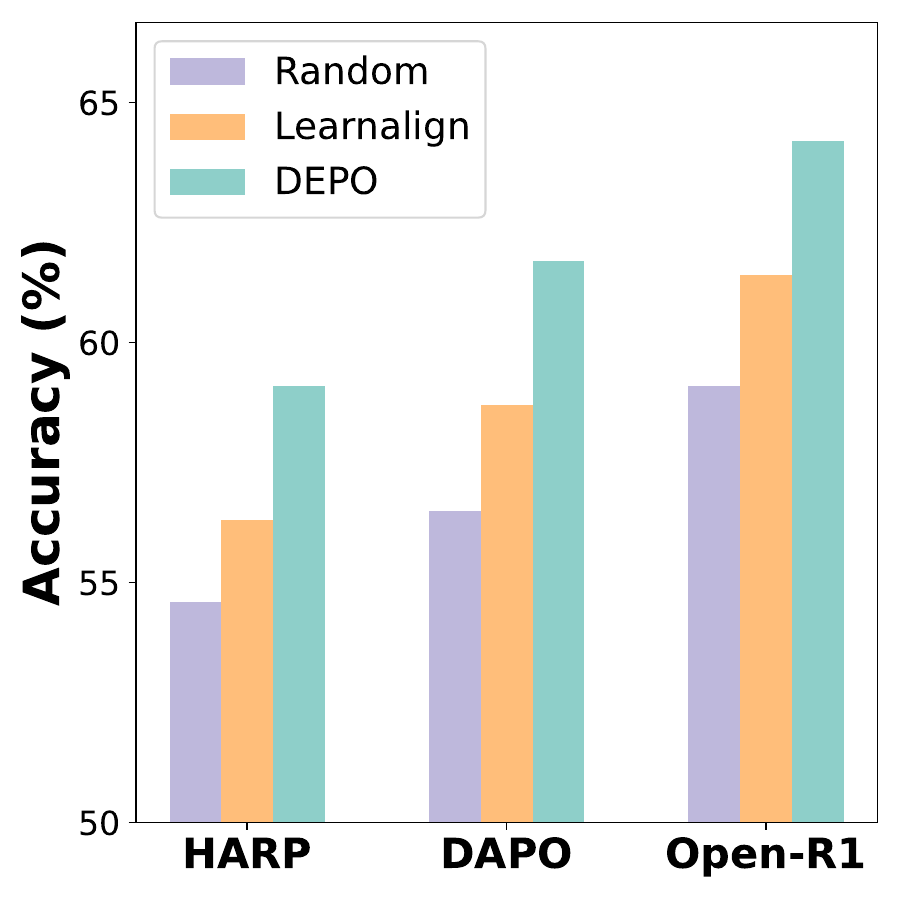}
            \caption{Offline Selection}
            \label{fig:dataset_offline}
        \end{subfigure}
        \begin{subfigure}[b]{0.49\linewidth}
            \centering
            \includegraphics[width=\linewidth]{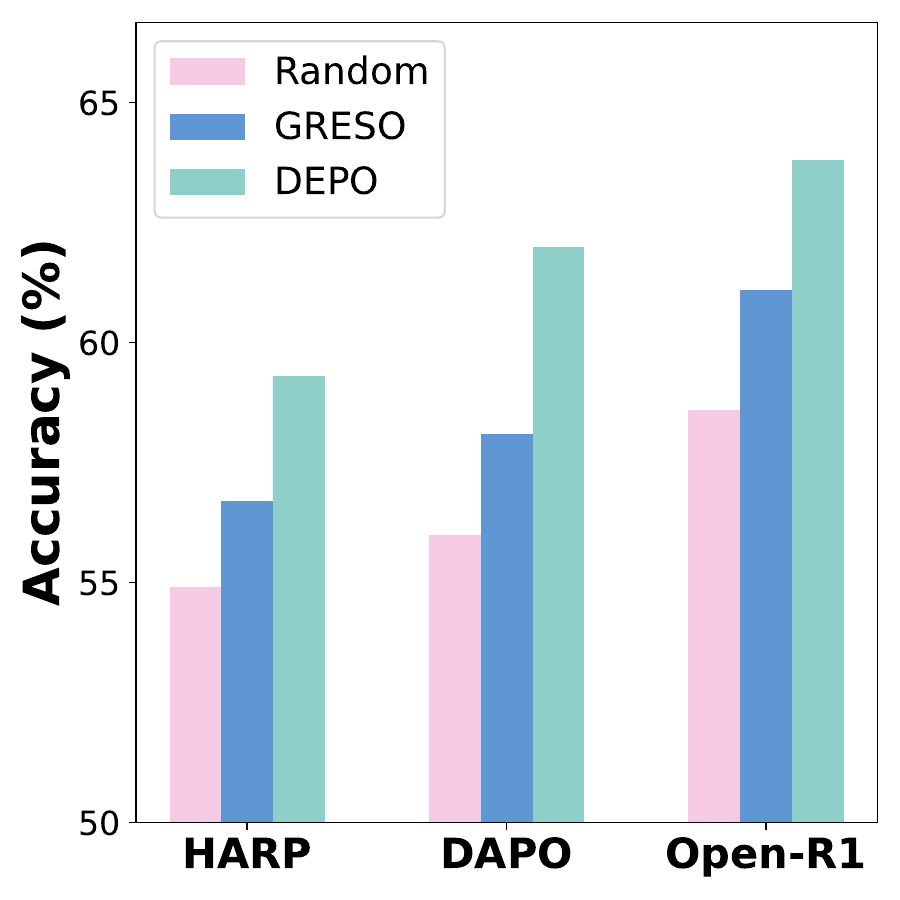}
            \caption{Online Selection}
            \label{fig:dataset_online}
        \end{subfigure}
        \caption{Different RLVR datasets.}
        \label{fig:RLVR_dataset}
    \end{minipage}
    \hfill
    \begin{minipage}[b]{0.49\linewidth}
        \centering
        \begin{subfigure}[b]{0.49\linewidth}
            \centering
            \includegraphics[width=\linewidth]{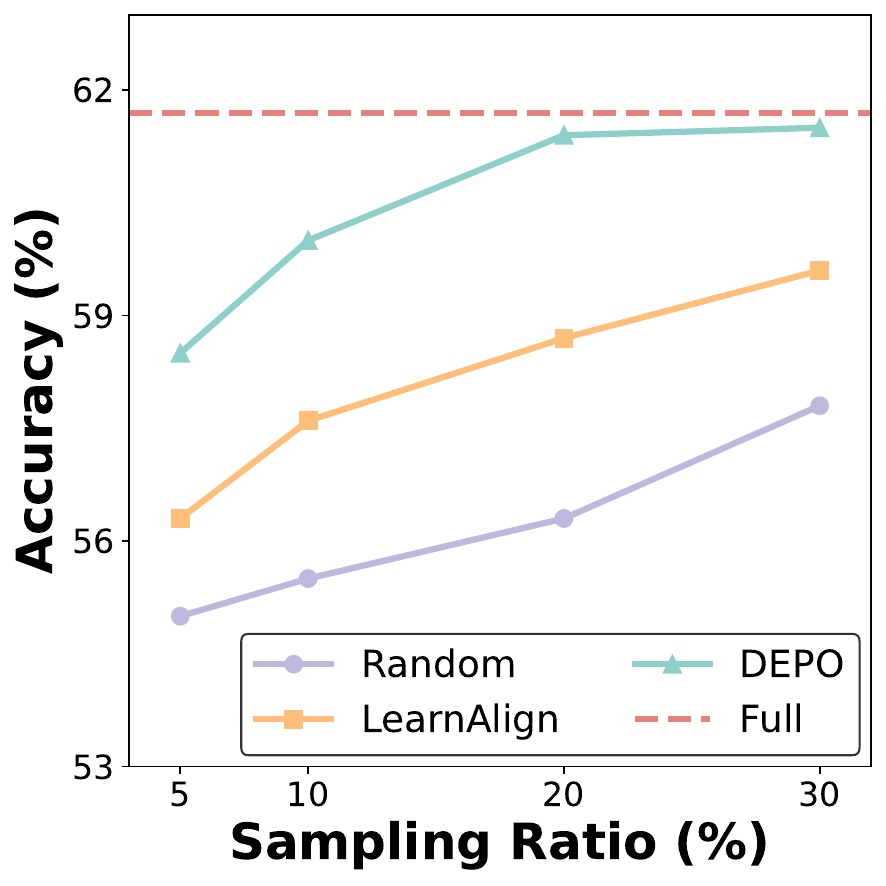}
            \caption{Offline Selection}
            \label{fig:ratio_offline}
        \end{subfigure}
        \begin{subfigure}[b]{0.49\linewidth}
            \centering
            \includegraphics[width=\linewidth]{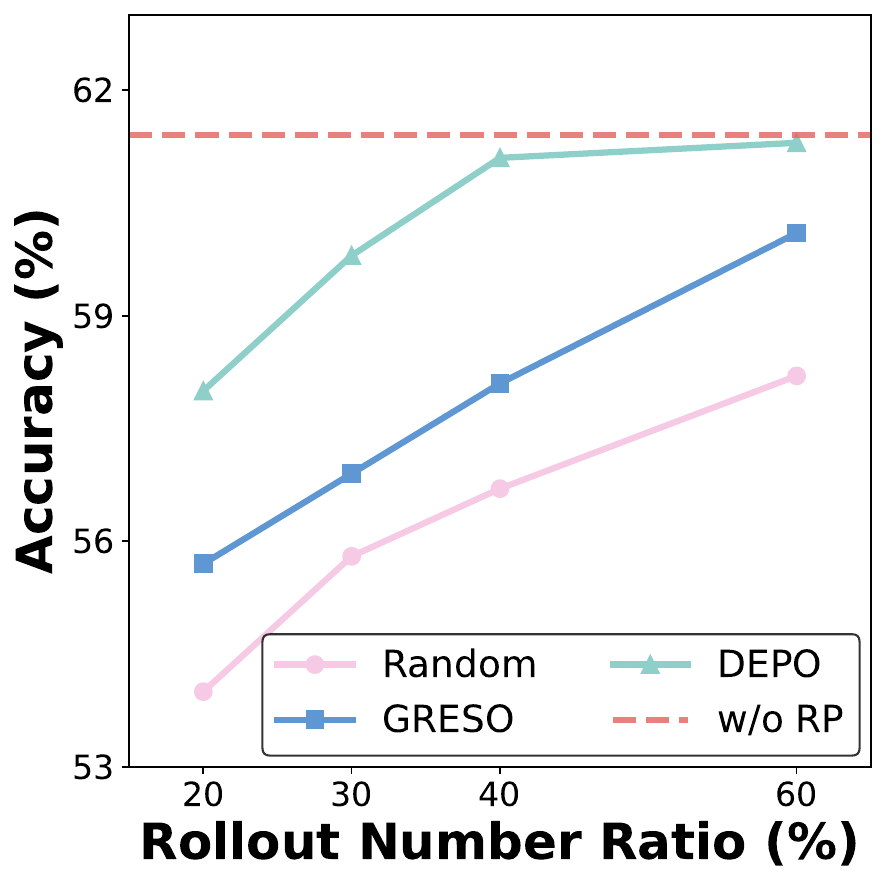}
            \caption{Online Selection}
            \label{fig:ratio_online}
        \end{subfigure}
        \caption{Different sampling ratios and rollout numbers.}
        \label{fig:sample_ratio}
    \end{minipage}
\end{figure}

\paratitle{\OURS performs well using different RLVR training datasets.}
To evaluate the effect of different training datasets, we conduct experiments using three commonly used RLVR datasets of varying sizes (\ie HARP~\citep{HARP}~(5k samples), DAPO~\citep{DAPO}~(17k samples), and Open-R1~\citep{openr1}~(30k samples)).
As illustrated in Figure~\ref{fig:RLVR_dataset}, performance improves consistently as the size of the training dataset increases, indicating that training on larger high-quality data can effectively enhance reasoning capabilities.
Besides, \OURS achieves strong performance using all three datasets, outperforming competitive baselines in both offline and online scenarios.
These results confirm that our approach is capable of improving data efficiency with different volumes of data.

\begin{figure}[h]
    \begin{subfigure}[b]{0.33\linewidth}
        \centering
        \includegraphics[width=\linewidth]{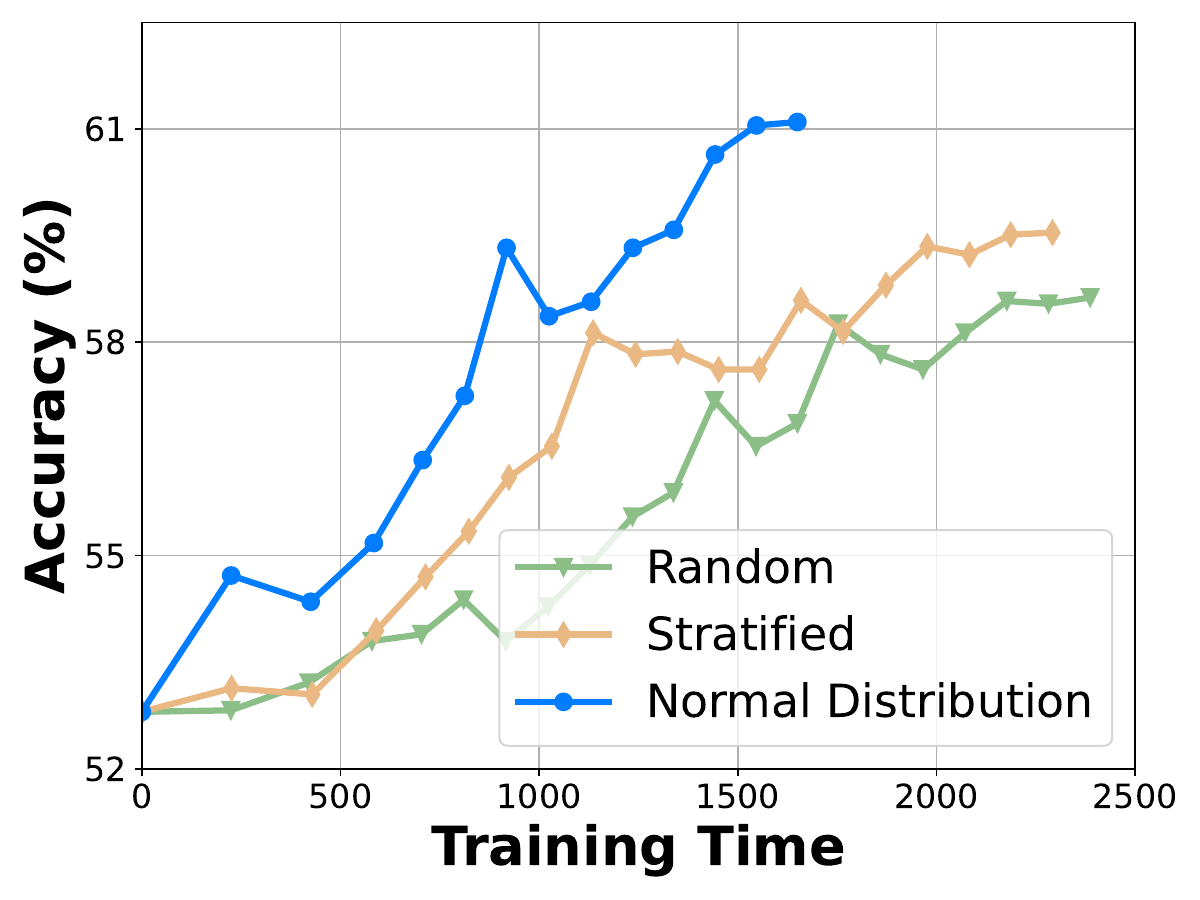}
        \caption{Different Sampling Strategy}
        \label{fig:sample_strategy}
    \end{subfigure}
    \begin{subfigure}[b]{0.33\linewidth}
        \centering
        \includegraphics[width=\linewidth]{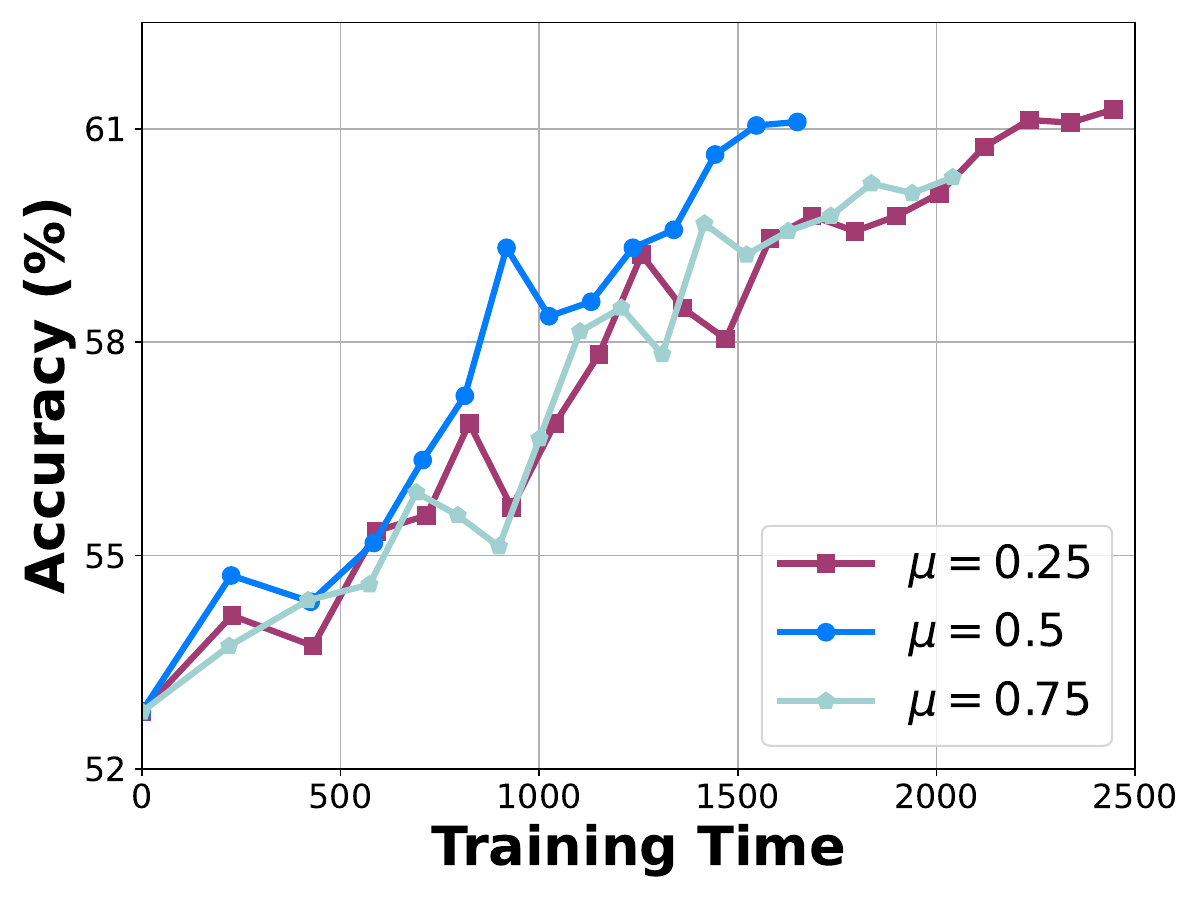}
        \caption{Different Mean}
        \label{fig:mean}
    \end{subfigure}
    \begin{subfigure}[b]{0.33\linewidth}
        \centering
        \includegraphics[width=\linewidth]{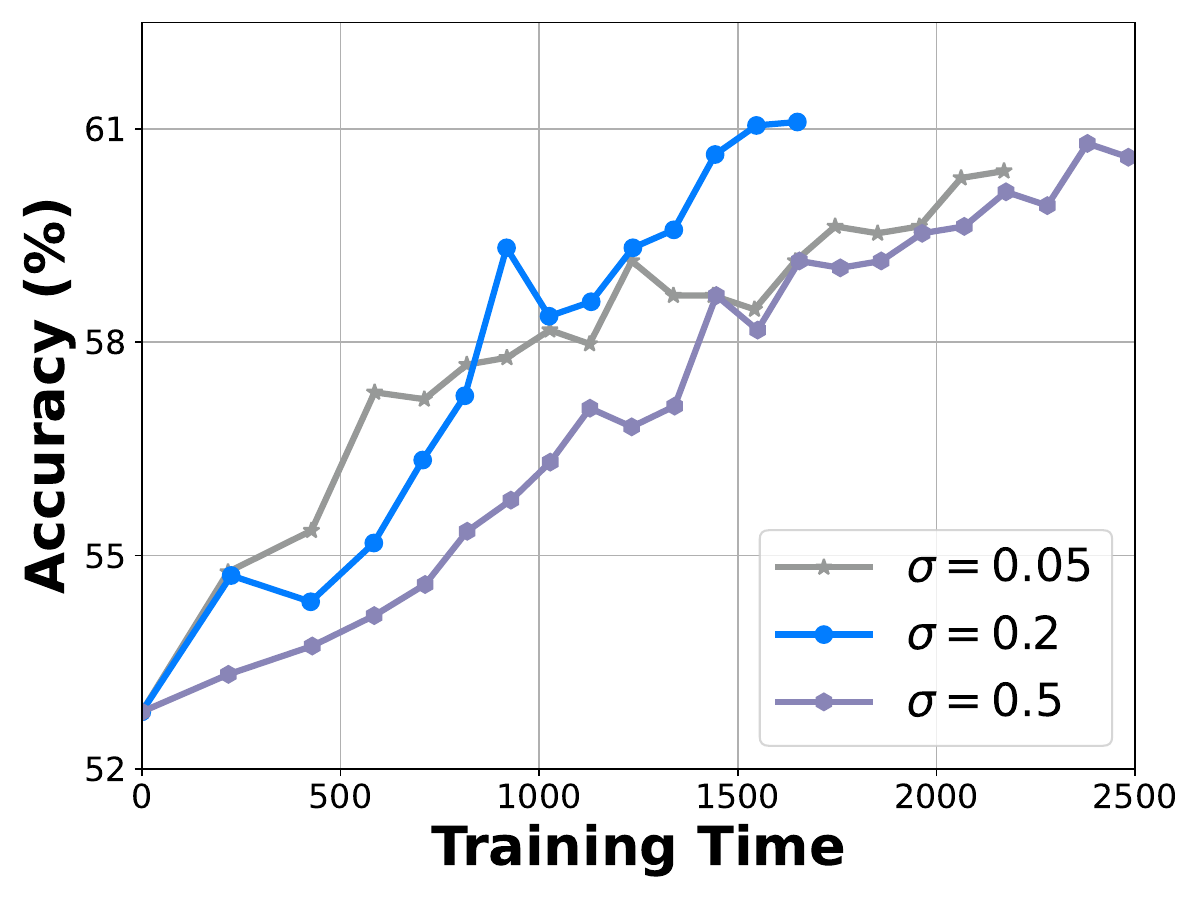}
        \caption{Different Variance}
        \label{fig:variance}
    \end{subfigure}
    \caption{Performance comparison across data of different difficulty levels.}
\label{fig:difficulty}
\end{figure}

\paratitle{\OURS performs well with different offline data selection sampling ratios.}
Figure~\ref{fig:ratio_offline} compares the performance of various offline data selection methods under different sampling ratios.
As illustrated, the results consistently follow the trend (\ie Random < LearnAlign < \OURS) across all ratios.
Notably, our method achieves performance competitive with full-dataset training even using only 20\% of the data.
This indicates that our approach identifies high-value samples, allowing the model to efficiently improve its reasoning capabilities with limited data.
Moreover, we observe that the performance of our method initially improves with increased sampling ratios, and it plateaus at around 20\%.
This also suggests that the dataset contains a substantial portion of redundant and low-value samples, which contribute minimally to the model's reasoning performance.

\paratitle{\OURS achieves the best performance across different rollout numbers.}
Figure~\ref{fig:ratio_online} presents the performance of various online data selection methods under various rollout ratios.
Our approach consistently outperforms all baselines under different rollout budgets.
Notably, selectively performing rollouts on samples with high explorability significantly improves training efficiency without substantially decreasing performance.
These results show the effectiveness of using explorability as a criterion to identify samples that contribute most to enhancing the model’s reasoning ability.

\paratitle{Sampling based on a normal distribution of difficulty can align the dataset's original difficulty with the current model's capability.}
To evaluate the impact of different data sampling strategies, we compare our normal distribution sampling approach with both random and stratified sampling.
Random sampling produces a difficulty distribution similar to the original one, with many samples being either entirely correct or entirely incorrect, resulting in a U-shaped distribution.
Stratified sampling selects samples from each difficulty level with equal probability.
As shown in Figure~\ref{fig:sample_strategy}, random sampling yields the lowest performance, followed by stratified sampling, while our method achieves the best results.
This is because extremely easy or hard samples offer limited training signals and contribute little to model improvement, and \OURS prioritizes samples of moderate difficulty, which are more informative and beneficial for RLVR training.

\paratitle{Training with medium-difficulty samples leads to rapid performance improvement, while the inclusion of challenging data enables higher final convergence performance.}
To further analyze the effect of difficulty-aware normal distribution sampling, we vary the mean and standard deviation of the normal distribution used for sampling, as shown in Figure~\ref{fig:mean} and Figure~\ref{fig:variance}.
Experimental results in Figure~\ref{fig:mean} indicate that using relatively easy samples (i.e., $\mu=0.75$) fails to achieve high convergence performance.
In contrast, using more challenging samples (\ie $\mu=0.25$) results in slightly better final performance, but requires more training time.
This suggests that medium-difficulty samples promote more efficient learning, which may be because they align well with the model's current capabilities.
Additionally, as shown in Figure~\ref{fig:variance}, we observe that a smaller standard deviation (\ie $\sigma=0.05$) leads to faster initial improvement but limits the final convergence performance.
Conversely, a larger standard deviation (\ie $\sigma=0.5$) slows down the learning progress but results in better final performance.
These findings indicate that medium-difficulty samples accelerate learning in the early stages, and incorporating a certain proportion of more challenging examples is crucial for achieving higher convergence.

\begin{wrapfigure}{r}{0.5\textwidth}
    \centering
    \begin{subfigure}[b]{0.48\linewidth}
        \centering
        \includegraphics[width=\linewidth]{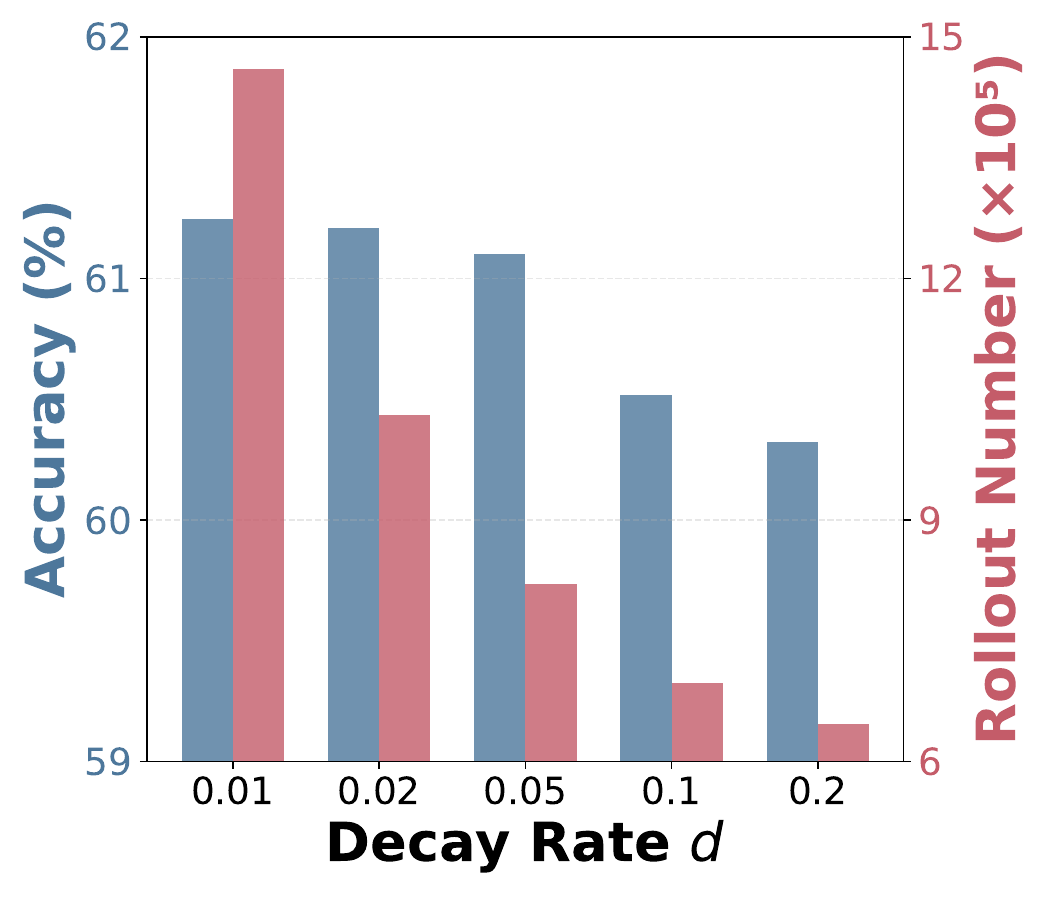}
        \caption{Decay Rate}
        \label{fig:decay_rate}
    \end{subfigure}
    \hfill
    \begin{subfigure}[b]{0.48\linewidth}
        \centering
        \includegraphics[width=\linewidth]{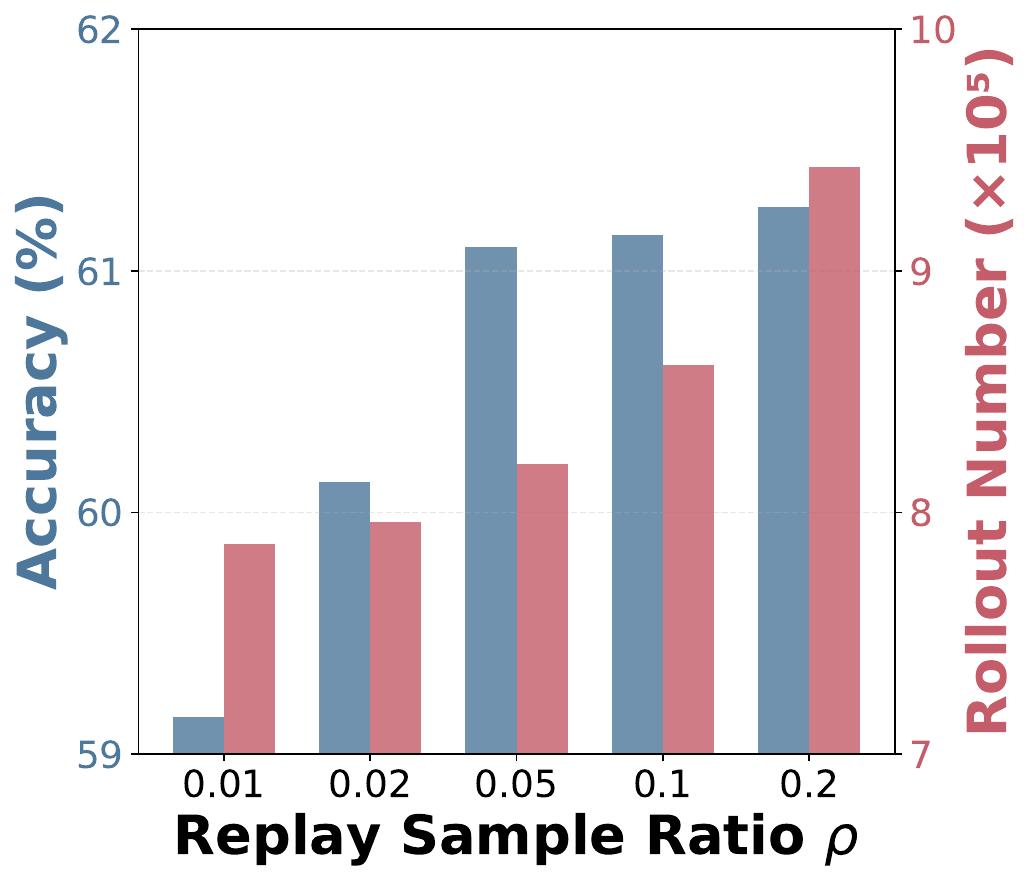}
        \caption{Replay Ratio}
        \label{fig:replay_ratio}
    \end{subfigure}
    \caption{Hyperparameter analysis}
    \label{fig:hyper_para1}
\end{wrapfigure}

\paratitle{Employing a suitable decay rate and replay sample ratio can significantly reduce the number of rollouts while maintaining comparable performance.}
In this paper, we adopt a linear decay strategy to gradually decrease the proportion of rollouts in each batch over training epochs.
As shown in Figure~\ref{fig:decay_rate}, setting the decay rate $d$ to 0.1 leads to a slight performance drop while substantially reducing rollout numbers.
If the decay rate is too high, many samples may not be sufficiently trained, leading to suboptimal final performance.
On the other hand, an excessively low decay rate would cause increased rollout numbers, thereby reducing training efficiency.
With respect to the replay ratio, setting it to 0.05 allows the model to achieve an optimal balance between final performance and training efficiency, as illustrated in Figure~\ref{fig:replay_ratio}. 
An excessively high replay ratio introduces unnecessary computational overhead due to redundant rollouts.
Conversely, if the replay ratio is too low, some challenging samples may not be adequately trained, hindering the model's reasoning abilities.

\section{Related Work}

\subsection{Reinforcement Learning with Verifiable Reward}

Reinforcement Learning with Verifiable Reward~(RLVR) has emerged as a promising paradigm for enhancing the complex reasoning capabilities of large language models~(LLMs)~\citep{LLM-survey}, particularly in domains such as mathematics~\citep{Agent-commute} and code generation~\citep{LiveCodeBench}.
The key advantage of this approach is its reward design, which relies solely on simple verification functions to provide binary rewards without requiring learned reward models.
DeepSeek-R1~\citep{Deepseek-r1} introduces the GRPO algorithm under the RLVR framework and demonstrates its effectiveness in significantly scaling the reasoning abilities of LLMs.
Building on GRPO, subsequent work have further advanced RLVR by refining various aspects, including loss functions~\citep{DR-GRPO,DAPO}, token-level entropy~\citep{80-20-rule}, advantage estimation~\citep{Advantage}, and hyperparameter~\citep{Hyperparameter}.
In this work, we focus on improving the data efficiency of RLVR to reduce computational costs while maintaining model performance.

\subsection{Data Efficiency for RLVR}

Data efficiency aims to enhance model performance by strategically selecting high-quality training samples.
Existing RLVR data selection approaches can be broadly categorized into offline and online strategies.
Offline data selection methods focus on identifying a high-quality subset of data prior to training.
Some studies select samples based on model reward trends~\citep{LIMR}, reward variance~\citep{1-shot-RLVR}, and gradient alignment~\citep{Learnalign}.
While effective, these methods require training the original or warmup dataset for several epochs for selection.
Another line of work~\citep{POLARIS} uses offline rollouts to estimate sample difficulty relative to the current model’s capabilities.
However, this approach requires computationally expensive rollouts across the entire dataset and fails to account for the diversity and influence of the training subsets.
On the other hand, online data selection methods aim to reduce training overhead by dynamically filtering samples during the training process.
These approaches target samples that contribute little to improving reasoning capabilities yet require costly rollouts.
GRESO~\citep{GRESO} employs a probabilistic filtering strategy to exclude samples with historical zero variance.
Although it improves efficiency, this approach only focuses on historical zero-variance samples and does not distinguish among other samples.
In this work, we enhance the data efficiency of RLVR by integrating optimized strategies for both offline and online data selection.
\section{Conclusion}

In this paper, we propose \OURS, a data-efficient policy optimization pipeline that integrates offline and online data selection strategies.
By first constructing a high-quality subset of training data that emphasizes diversity, influence, and appropriate difficulty, and then dynamically filtering rollouts during online training based on sample-level explorability, our approach significantly reduces both data volume and computational costs while maintaining comparable performance.
Extensive experiments across multiple reasoning benchmarks and LLMs demonstrate that \OURS consistently outperforms competitive baselines in both offline and online settings.
We hope our work inspires future research toward developing more data-efficient methods to accelerate RL for LLMs.

\bibliographystyle{assets/plainnat}
\bibliography{main}

\begin{thebibliography}{38}
\providecommand{\natexlab}[1]{#1}
\providecommand{\url}[1]{\texttt{#1}}
\expandafter\ifx\csname urlstyle\endcsname\relax
  \providecommand{\doi}[1]{doi: #1}\else
  \providecommand{\doi}{doi: \begingroup \urlstyle{rm}\Url}\fi

\bibitem[An et~al.(2025)An, Xie, Li, Li, Zhang, Gong, Zhong, Xu, Qiu, Wang, and Kong]{POLARIS}
Chenxin An, Zhihui Xie, Xiaonan Li, Lei Li, Jun Zhang, Shansan Gong, Ming Zhong, Jingjing Xu, Xipeng Qiu, Mingxuan Wang, and Lingpeng Kong.
\newblock Polaris: A post-training recipe for scaling reinforcement learning on advanced reasoning models, 2025.
\newblock \url{https://hkunlp.github.io/blog/2025/Polaris}.

\bibitem[Ankner et~al.(2025)Ankner, Blakeney, Sreenivasan, Marion, Leavitt, and Paul]{PPL-Middle}
Zachary Ankner, Cody Blakeney, Kartik Sreenivasan, Max Marion, Matthew~L. Leavitt, and Mansheej Paul.
\newblock Perplexed by perplexity: Perplexity-based data pruning with small reference models.
\newblock In \emph{{ICLR}}. OpenReview.net, 2025.

\bibitem[Brin and Page(1998)]{PageRank}
Sergey Brin and Lawrence Page.
\newblock The anatomy of a large-scale hypertextual web search engine.
\newblock \emph{Comput. Networks}, 30\penalty0 (1-7):\penalty0 107--117, 1998.

\bibitem[Chen et~al.(2018)Chen, Zhang, and Zhou]{DPP2}
Laming Chen, Guoxin Zhang, and Eric Zhou.
\newblock Fast greedy {MAP} inference for determinantal point process to improve recommendation diversity.
\newblock In \emph{NeurIPS}, pages 5627--5638, 2018.

\bibitem[Cheng et~al.(2025)Cheng, Huang, Zhu, Dai, Zhao, Zhang, and Wei]{Advantage}
Daixuan Cheng, Shaohan Huang, Xuekai Zhu, Bo~Dai, Wayne~Xin Zhao, Zhenliang Zhang, and Furu Wei.
\newblock Reasoning with exploration: An entropy perspective.
\newblock \emph{CoRR}, abs/2506.14758, 2025.

\bibitem[DeepSeek{-}AI et~al.(2025)DeepSeek{-}AI, Guo, Yang, Zhang, Song, Zhang, Xu, Zhu, Ma, Wang, Bi, Zhang, Yu, Wu, Wu, Gou, Shao, Li, Gao, Liu, Xue, Wang, Wu, Feng, Lu, Zhao, Deng, Zhang, Ruan, Dai, Chen, Ji, Li, Lin, Dai, Luo, Hao, Chen, Li, Zhang, Bao, Xu, Wang, Ding, Xin, Gao, Qu, Li, Guo, Li, Wang, Chen, Yuan, Qiu, Li, Cai, Ni, Liang, Chen, Dong, Hu, Gao, Guan, Huang, Yu, Wang, Zhang, Zhao, Wang, Zhang, Xu, Xia, Zhang, Zhang, Tang, Li, Wang, Li, Tian, Huang, Zhang, Wang, Chen, Du, Ge, Zhang, Pan, Wang, Chen, Jin, Chen, Lu, Zhou, Chen, Ye, Wang, Yu, Zhou, Pan, and Li]{Deepseek-r1}
DeepSeek{-}AI, Daya Guo, Dejian Yang, Haowei Zhang, Junxiao Song, Ruoyu Zhang, Runxin Xu, Qihao Zhu, Shirong Ma, Peiyi Wang, Xiao Bi, Xiaokang Zhang, Xingkai Yu, Yu~Wu, Z.~F. Wu, Zhibin Gou, Zhihong Shao, Zhuoshu Li, Ziyi Gao, Aixin Liu, Bing Xue, Bingxuan Wang, Bochao Wu, Bei Feng, Chengda Lu, Chenggang Zhao, Chengqi Deng, Chenyu Zhang, Chong Ruan, Damai Dai, Deli Chen, Dongjie Ji, Erhang Li, Fangyun Lin, Fucong Dai, Fuli Luo, Guangbo Hao, Guanting Chen, Guowei Li, H.~Zhang, Han Bao, Hanwei Xu, Haocheng Wang, Honghui Ding, Huajian Xin, Huazuo Gao, Hui Qu, Hui Li, Jianzhong Guo, Jiashi Li, Jiawei Wang, Jingchang Chen, Jingyang Yuan, Junjie Qiu, Junlong Li, J.~L. Cai, Jiaqi Ni, Jian Liang, Jin Chen, Kai Dong, Kai Hu, Kaige Gao, Kang Guan, Kexin Huang, Kuai Yu, Lean Wang, Lecong Zhang, Liang Zhao, Litong Wang, Liyue Zhang, Lei Xu, Leyi Xia, Mingchuan Zhang, Minghua Zhang, Minghui Tang, Meng Li, Miaojun Wang, Mingming Li, Ning Tian, Panpan Huang, Peng Zhang, Qiancheng Wang, Qinyu Chen, Qiushi Du, Ruiqi Ge,
  Ruisong Zhang, Ruizhe Pan, Runji Wang, R.~J. Chen, R.~L. Jin, Ruyi Chen, Shanghao Lu, Shangyan Zhou, Shanhuang Chen, Shengfeng Ye, Shiyu Wang, Shuiping Yu, Shunfeng Zhou, Shuting Pan, and S.~S. Li.
\newblock Deepseek-r1: Incentivizing reasoning capability in llms via reinforcement learning.
\newblock \emph{CoRR}, abs/2501.12948, 2025.

\bibitem[Face(2025)]{openr1}
Hugging Face.
\newblock Open r1: A fully open reproduction of deepseek-r1, January 2025.
\newblock \url{https://github.com/huggingface/open-r1}.

\bibitem[Hendel et~al.(2023)Hendel, Geva, and Globerson]{task-vector}
Roee Hendel, Mor Geva, and Amir Globerson.
\newblock In-context learning creates task vectors.
\newblock In \emph{{EMNLP} (Findings)}, pages 9318--9333. Association for Computational Linguistics, 2023.

\bibitem[Hendrycks et~al.(2021)Hendrycks, Burns, Kadavath, Arora, Basart, Tang, Song, and Steinhardt]{Math500}
Dan Hendrycks, Collin Burns, Saurav Kadavath, Akul Arora, Steven Basart, Eric Tang, Dawn Song, and Jacob Steinhardt.
\newblock Measuring mathematical problem solving with the {MATH} dataset.
\newblock In \emph{NeurIPS Datasets and Benchmarks}, 2021.

\bibitem[Jaech et~al.(2024)Jaech, Kalai, Lerer, Richardson, El{-}Kishky, Low, Helyar, Madry, Beutel, Carney, Iftimie, Karpenko, Passos, Neitz, Prokofiev, Wei, Tam, Bennett, Kumar, Saraiva, Vallone, Duberstein, Kondrich, Mishchenko, Applebaum, Jiang, Nair, Zoph, Ghorbani, Rossen, Sokolowsky, Barak, McGrew, Minaiev, Hao, Baker, Houghton, McKinzie, Eastman, Lugaresi, Bassin, Hudson, Li, de~Bourcy, Voss, Shen, Zhang, Koch, Orsinger, Hesse, Fischer, Chan, Roberts, Kappler, Levy, Selsam, Dohan, Farhi, Mely, Robinson, Tsipras, Li, Oprica, Freeman, Zhang, Wong, Proehl, Cheung, Mitchell, Wallace, Ritter, Mays, Wang, Such, Raso, Leoni, Tsimpourlas, Song, von Lohmann, Sulit, Salmon, Parascandolo, Chabot, Zhao, Brockman, Leclerc, Salman, Bao, Sheng, Andrin, Bagherinezhad, Ren, Lightman, Chung, Kivlichan, O'Connell, Osband, Gilaberte, and Akkaya]{OpenAI-o1}
Aaron Jaech, Adam Kalai, Adam Lerer, Adam Richardson, Ahmed El{-}Kishky, Aiden Low, Alec Helyar, Aleksander Madry, Alex Beutel, Alex Carney, Alex Iftimie, Alex Karpenko, Alex~Tachard Passos, Alexander Neitz, Alexander Prokofiev, Alexander Wei, Allison Tam, Ally Bennett, Ananya Kumar, Andre Saraiva, Andrea Vallone, Andrew Duberstein, Andrew Kondrich, Andrey Mishchenko, Andy Applebaum, Angela Jiang, Ashvin Nair, Barret Zoph, Behrooz Ghorbani, Ben Rossen, Benjamin Sokolowsky, Boaz Barak, Bob McGrew, Borys Minaiev, Botao Hao, Bowen Baker, Brandon Houghton, Brandon McKinzie, Brydon Eastman, Camillo Lugaresi, Cary Bassin, Cary Hudson, Chak~Ming Li, Charles de~Bourcy, Chelsea Voss, Chen Shen, Chong Zhang, Chris Koch, Chris Orsinger, Christopher Hesse, Claudia Fischer, Clive Chan, Dan Roberts, Daniel Kappler, Daniel Levy, Daniel Selsam, David Dohan, David Farhi, David Mely, David Robinson, Dimitris Tsipras, Doug Li, Dragos Oprica, Eben Freeman, Eddie Zhang, Edmund Wong, Elizabeth Proehl, Enoch Cheung, Eric Mitchell,
  Eric Wallace, Erik Ritter, Evan Mays, Fan Wang, Felipe~Petroski Such, Filippo Raso, Florencia Leoni, Foivos Tsimpourlas, Francis Song, Fred von Lohmann, Freddie Sulit, Geoff Salmon, Giambattista Parascandolo, Gildas Chabot, Grace Zhao, Greg Brockman, Guillaume Leclerc, Hadi Salman, Haiming Bao, Hao Sheng, Hart Andrin, Hessam Bagherinezhad, Hongyu Ren, Hunter Lightman, Hyung~Won Chung, Ian Kivlichan, Ian O'Connell, Ian Osband, Ignasi~Clavera Gilaberte, and Ilge Akkaya.
\newblock Openai o1 system card.
\newblock \emph{CoRR}, abs/2412.16720, 2024.

\bibitem[Jain et~al.(2025)Jain, Han, Gu, Li, Yan, Zhang, Wang, Solar{-}Lezama, Sen, and Stoica]{LiveCodeBench}
Naman Jain, King Han, Alex Gu, Wen{-}Ding Li, Fanjia Yan, Tianjun Zhang, Sida Wang, Armando Solar{-}Lezama, Koushik Sen, and Ion Stoica.
\newblock Livecodebench: Holistic and contamination free evaluation of large language models for code.
\newblock In \emph{{ICLR}}. OpenReview.net, 2025.

\bibitem[Kulesza and Taskar(2012{\natexlab{a}})]{DPP}
Alex Kulesza and Ben Taskar.
\newblock Determinantal point processes for machine learning.
\newblock \emph{CoRR}, abs/1207.6083, 2012{\natexlab{a}}.

\bibitem[Kulesza and Taskar(2012{\natexlab{b}})]{DPP1}
Alex Kulesza and Ben Taskar.
\newblock Determinantal point processes for machine learning.
\newblock \emph{Found. Trends Mach. Learn.}, 5\penalty0 (2-3):\penalty0 123--286, 2012{\natexlab{b}}.

\bibitem[Kwon et~al.(2023)Kwon, Li, Zhuang, Sheng, Zheng, Yu, Gonzalez, Zhang, and Stoica]{vllm}
Woosuk Kwon, Zhuohan Li, Siyuan Zhuang, Ying Sheng, Lianmin Zheng, Cody~Hao Yu, Joseph Gonzalez, Hao Zhang, and Ion Stoica.
\newblock Efficient memory management for large language model serving with pagedattention.
\newblock In \emph{{SOSP}}, pages 611--626. {ACM}, 2023.

\bibitem[Lauren{\c{c}}on et~al.(2022)Lauren{\c{c}}on, Saulnier, Wang, Akiki, del Moral, Scao, von Werra, Mou, Ponferrada, Nguyen, Frohberg, Sasko, Lhoest, McMillan{-}Major, Dupont, Biderman, Rogers, Allal, Toni, Pistilli, Nguyen, Nikpoor, Masoud, Colombo, de~la Rosa, Villegas, Thrush, Longpre, Nagel, Weber, Mu{\~{n}}oz, Zhu, van Strien, Alyafeai, Almubarak, Vu, Gonzalez{-}Dios, Soroa, Lo, Dey, Suarez, Gokaslan, Bose, Adelani, Phan, Tran, Yu, Pai, Chim, Lepercq, Ilic, Mitchell, Luccioni, and Jernite]{PPL-Top}
Hugo Lauren{\c{c}}on, Lucile Saulnier, Thomas Wang, Christopher Akiki, Albert~Villanova del Moral, Teven~Le Scao, Leandro von Werra, Chenghao Mou, Eduardo~Gonz{\'{a}}lez Ponferrada, Huu Nguyen, J{\"{o}}rg Frohberg, Mario Sasko, Quentin Lhoest, Angelina McMillan{-}Major, G{\'{e}}rard Dupont, Stella Biderman, Anna Rogers, Loubna~Ben Allal, Francesco~De Toni, Giada Pistilli, Olivier Nguyen, Somaieh Nikpoor, Maraim Masoud, Pierre Colombo, Javier de~la Rosa, Paulo Villegas, Tristan Thrush, Shayne Longpre, Sebastian Nagel, Leon Weber, Manuel Mu{\~{n}}oz, Jian Zhu, Daniel van Strien, Zaid Alyafeai, Khalid Almubarak, Minh~Chien Vu, Itziar Gonzalez{-}Dios, Aitor Soroa, Kyle Lo, Manan Dey, Pedro~Ortiz Suarez, Aaron Gokaslan, Shamik Bose, David~Ifeoluwa Adelani, Long Phan, Hieu Tran, Ian Yu, Suhas Pai, Jenny Chim, Violette Lepercq, Suzana Ilic, Margaret Mitchell, Alexandra~Sasha Luccioni, and Yacine Jernite.
\newblock The bigscience {ROOTS} corpus: {A} 1.6tb composite multilingual dataset.
\newblock In \emph{NeurIPS}, 2022.

\bibitem[Li et~al.(2025{\natexlab{a}})Li, Li, Yang, Zhang, Chen, Xia, Liu, and Peng]{Learnalign}
Shikun Li, Shipeng Li, Zhiqin Yang, Xinghua Zhang, Gaode Chen, Xiaobo Xia, Hengyu Liu, and Zhe Peng.
\newblock Learnalign: Reasoning data selection for reinforcement learning in large language models based on improved gradient alignment.
\newblock \emph{CoRR}, abs/2506.11480, 2025{\natexlab{a}}.

\bibitem[Li et~al.(2025{\natexlab{b}})Li, Zou, and Liu]{LIMR}
Xuefeng Li, Haoyang Zou, and Pengfei Liu.
\newblock {LIMR:} less is more for {RL} scaling.
\newblock \emph{CoRR}, abs/2502.11886, 2025{\natexlab{b}}.

\bibitem[Liu et~al.(2024)Liu, Ye, Xing, and Zou]{in-context-vector}
Sheng Liu, Haotian Ye, Lei Xing, and James~Y. Zou.
\newblock In-context vectors: Making in context learning more effective and controllable through latent space steering.
\newblock In \emph{{ICML}}. OpenReview.net, 2024.

\bibitem[Liu et~al.(2025{\natexlab{a}})Liu, Chen, Li, Qi, Pang, Du, Lee, and Lin]{DR-GRPO}
Zichen Liu, Changyu Chen, Wenjun Li, Penghui Qi, Tianyu Pang, Chao Du, Wee~Sun Lee, and Min Lin.
\newblock Understanding r1-zero-like training: {A} critical perspective.
\newblock \emph{CoRR}, abs/2503.20783, 2025{\natexlab{a}}.

\bibitem[Liu et~al.(2025{\natexlab{b}})Liu, Liu, He, Wang, Liu, Pan, Hu, Xiong, Huang, Hu, Huang, Yang, Wang, Su, and Zheng]{Hyperparameter}
Zihe Liu, Jiashun Liu, Yancheng He, Weixun Wang, Jiaheng Liu, Ling Pan, Xinyu Hu, Shaopan Xiong, Ju~Huang, Jian Hu, Shengyi Huang, Siran Yang, Jiamang Wang, Wenbo Su, and Bo~Zheng.
\newblock Part i: Tricks or traps? a deep dive into rl for llm reasoning, 2025{\natexlab{b}}.
\newblock \url{https://arxiv.org/abs/2508.08221}.

\bibitem[Loshchilov and Hutter(2019)]{AdamW}
Ilya Loshchilov and Frank Hutter.
\newblock Decoupled weight decay regularization.
\newblock In \emph{{ICLR} (Poster)}. OpenReview.net, 2019.

\bibitem[Ouyang et~al.(2022)Ouyang, Wu, Jiang, Almeida, Wainwright, Mishkin, Zhang, Agarwal, Slama, Ray, Schulman, Hilton, Kelton, Miller, Simens, Askell, Welinder, Christiano, Leike, and Lowe]{PPO}
Long Ouyang, Jeffrey Wu, Xu~Jiang, Diogo Almeida, Carroll~L. Wainwright, Pamela Mishkin, Chong Zhang, Sandhini Agarwal, Katarina Slama, Alex Ray, John Schulman, Jacob Hilton, Fraser Kelton, Luke Miller, Maddie Simens, Amanda Askell, Peter Welinder, Paul~F. Christiano, Jan Leike, and Ryan Lowe.
\newblock Training language models to follow instructions with human feedback.
\newblock In \emph{NeurIPS}, 2022.

\bibitem[Rein et~al.(2023)Rein, Hou, Stickland, Petty, Pang, Dirani, Michael, and Bowman]{GPQA}
David Rein, Betty~Li Hou, Asa~Cooper Stickland, Jackson Petty, Richard~Yuanzhe Pang, Julien Dirani, Julian Michael, and Samuel~R. Bowman.
\newblock {GPQA:} {A} graduate-level google-proof q{\&}a benchmark.
\newblock \emph{CoRR}, abs/2311.12022, 2023.

\bibitem[Shao et~al.(2024)Shao, Wang, Zhu, Xu, Song, Zhang, Li, Wu, and Guo]{GRPO}
Zhihong Shao, Peiyi Wang, Qihao Zhu, Runxin Xu, Junxiao Song, Mingchuan Zhang, Y.~K. Li, Y.~Wu, and Daya Guo.
\newblock Deepseekmath: Pushing the limits of mathematical reasoning in open language models.
\newblock \emph{CoRR}, abs/2402.03300, 2024.

\bibitem[Sheng et~al.(2025)Sheng, Zhang, Ye, Wu, Zhang, Zhang, Peng, Lin, and Wu]{verl}
Guangming Sheng, Chi Zhang, Zilingfeng Ye, Xibin Wu, Wang Zhang, Ru~Zhang, Yanghua Peng, Haibin Lin, and Chuan Wu.
\newblock Hybridflow: {A} flexible and efficient {RLHF} framework.
\newblock In \emph{EuroSys}, pages 1279--1297. {ACM}, 2025.

\bibitem[Stolfo et~al.(2025)Stolfo, Balachandran, Yousefi, Horvitz, and Nushi]{if-vector}
Alessandro Stolfo, Vidhisha Balachandran, Safoora Yousefi, Eric Horvitz, and Besmira Nushi.
\newblock Improving instruction-following in language models through activation steering.
\newblock In \emph{{ICLR}}. OpenReview.net, 2025.

\bibitem[Tang et~al.(2025)Tang, Wang, Lv, Min, Zhao, Hu, Liu, and Zhang]{long-cot-vector}
Xinyu Tang, Xiaolei Wang, Zhihao Lv, Yingqian Min, Xin Zhao, Binbin Hu, Ziqi Liu, and Zhiqiang Zhang.
\newblock Unlocking general long chain-of-thought reasoning capabilities of large language models via representation engineering.
\newblock In \emph{{ACL} {(1)}}, pages 6832--6849. Association for Computational Linguistics, 2025.

\bibitem[Team et~al.(2025)Team, Du, Gao, Xing, Jiang, Chen, Li, Xiao, Du, Liao, Tang, Wang, Zhang, Yuan, Lu, Tang, Sung, Wei, Lai, Guo, Zhu, Ding, Hu, Yang, Zhang, Yao, Zhao, Lu, Li, Yu, Gao, Zheng, Yuan, Chen, Guo, Su, Wang, Zhao, Zhang, Liu, Yan, Wu, Shi, Ye, Yu, Dong, Zhang, Ma, Pan, Gong, Liu, Ma, Wei, Cao, Huang, Jiang, Gao, Xiong, He, Huang, Wu, He, Wei, Jia, Wu, Xu, Zu, Zhou, Pan, Charles, Li, Hu, Liu, Chen, Wang, Liu, Qin, Liu, Yang, Bao, Du, Wu, Wang, Zhou, Wang, Li, Zhu, Zhang, Wang, Yang, Huang, Huang, Xu, and Yang]{Kimi-1.5}
Kimi Team, Angang Du, Bofei Gao, Bowei Xing, Changjiu Jiang, Cheng Chen, Cheng Li, Chenjun Xiao, Chenzhuang Du, Chonghua Liao, Chuning Tang, Congcong Wang, Dehao Zhang, Enming Yuan, Enzhe Lu, Fengxiang Tang, Flood Sung, Guangda Wei, Guokun Lai, Haiqing Guo, Han Zhu, Hao Ding, Hao Hu, Hao Yang, Hao Zhang, Haotian Yao, Haotian Zhao, Haoyu Lu, Haoze Li, Haozhen Yu, Hongcheng Gao, Huabin Zheng, Huan Yuan, Jia Chen, Jianhang Guo, Jianlin Su, Jianzhou Wang, Jie Zhao, Jin Zhang, Jingyuan Liu, Junjie Yan, Junyan Wu, Lidong Shi, Ling Ye, Longhui Yu, Mengnan Dong, Neo Zhang, Ningchen Ma, Qiwei Pan, Qucheng Gong, Shaowei Liu, Shengling Ma, Shupeng Wei, Sihan Cao, Siying Huang, Tao Jiang, Weihao Gao, Weimin Xiong, Weiran He, Weixiao Huang, Wenhao Wu, Wenyang He, Xianghui Wei, Xianqing Jia, Xingzhe Wu, Xinran Xu, Xinxing Zu, Xinyu Zhou, Xuehai Pan, Y.~Charles, Yang Li, Yangyang Hu, Yangyang Liu, Yanru Chen, Yejie Wang, Yibo Liu, Yidao Qin, Yifeng Liu, Ying Yang, Yiping Bao, Yulun Du, Yuxin Wu, Yuzhi Wang, Zaida Zhou,
  Zhaoji Wang, Zhaowei Li, Zhen Zhu, Zheng Zhang, Zhexu Wang, Zhilin Yang, Zhiqi Huang, Zihao Huang, Ziyao Xu, and Zonghan Yang.
\newblock Kimi k1.5: Scaling reinforcement learning with llms.
\newblock \emph{CoRR}, abs/2501.12599, 2025.

\bibitem[Wang et~al.(2025{\natexlab{a}})Wang, Yu, Gao, Zheng, Liu, Lu, Dang, Chen, Yang, Zhang, Liu, Yang, Zhao, Yue, Song, Yu, Huang, and Lin]{80-20-rule}
Shenzhi Wang, Le~Yu, Chang Gao, Chujie Zheng, Shixuan Liu, Rui Lu, Kai Dang, Xionghui Chen, Jianxin Yang, Zhenru Zhang, Yuqiong Liu, An~Yang, Andrew Zhao, Yang Yue, Shiji Song, Bowen Yu, Gao Huang, and Junyang Lin.
\newblock Beyond the 80/20 rule: High-entropy minority tokens drive effective reinforcement learning for {LLM} reasoning.
\newblock \emph{CoRR}, abs/2506.01939, 2025{\natexlab{a}}.

\bibitem[Wang et~al.(2025{\natexlab{b}})Wang, Yang, Zeng, Ren, Liu, Peng, Cheng, He, Wang, Gao, Chen, Wang, Du, and Shen]{1-shot-RLVR}
Yiping Wang, Qing Yang, Zhiyuan Zeng, Liliang Ren, Lucas Liu, Baolin Peng, Hao Cheng, Xuehai He, Kuan Wang, Jianfeng Gao, Weizhu Chen, Shuohang Wang, Simon~Shaolei Du, and Yelong Shen.
\newblock Reinforcement learning for reasoning in large language models with one training example.
\newblock \emph{CoRR}, abs/2504.20571, 2025{\natexlab{b}}.

\bibitem[Yang et~al.(2024)Yang, Zhang, Hui, Gao, Yu, Li, Liu, Tu, Zhou, Lin, Lu, Xue, Lin, Liu, Ren, and Zhang]{Qwen2.5-Math-7B}
An~Yang, Beichen Zhang, Binyuan Hui, Bofei Gao, Bowen Yu, Chengpeng Li, Dayiheng Liu, Jianhong Tu, Jingren Zhou, Junyang Lin, Keming Lu, Mingfeng Xue, Runji Lin, Tianyu Liu, Xingzhang Ren, and Zhenru Zhang.
\newblock Qwen2.5-math technical report: Toward mathematical expert model via self-improvement.
\newblock \emph{CoRR}, abs/2409.12122, 2024.

\bibitem[Yu et~al.(2025)Yu, Zhang, Zhu, Yuan, Zuo, Yue, Fan, Liu, Liu, Liu, Lin, Lin, Ma, Sheng, Tong, Zhang, Zhang, Zhang, Zhu, Zhu, Chen, Chen, Wang, Yu, Dai, Song, Wei, Zhou, Liu, Ma, Zhang, Yan, Qiao, Wu, and Wang]{DAPO}
Qiying Yu, Zheng Zhang, Ruofei Zhu, Yufeng Yuan, Xiaochen Zuo, Yu~Yue, Tiantian Fan, Gaohong Liu, Lingjun Liu, Xin Liu, Haibin Lin, Zhiqi Lin, Bole Ma, Guangming Sheng, Yuxuan Tong, Chi Zhang, Mofan Zhang, Wang Zhang, Hang Zhu, Jinhua Zhu, Jiaze Chen, Jiangjie Chen, Chengyi Wang, Hongli Yu, Weinan Dai, Yuxuan Song, Xiangpeng Wei, Hao Zhou, Jingjing Liu, Wei{-}Ying Ma, Ya{-}Qin Zhang, Lin Yan, Mu~Qiao, Yonghui Wu, and Mingxuan Wang.
\newblock {DAPO:} an open-source {LLM} reinforcement learning system at scale.
\newblock \emph{CoRR}, abs/2503.14476, 2025.

\bibitem[Yue et~al.(2024)Yue, Madaan, Moskovitz, Strouse, and Singh]{HARP}
Albert~S. Yue, Lovish Madaan, Ted Moskovitz, DJ~Strouse, and Aaditya~K. Singh.
\newblock {HARP:} {A} challenging human-annotated math reasoning benchmark.
\newblock \emph{CoRR}, abs/2412.08819, 2024.

\bibitem[Yue et~al.(2025)Yue, Yuan, Yu, Zuo, Zhu, Xu, Chen, Wang, Fan, Du, Wei, Yu, Liu, Liu, Liu, Lin, Lin, Ma, Zhang, Zhang, Zhang, Zhu, Zhang, Liu, Wang, Wu, and Yan]{VAPO}
Yu~Yue, Yufeng Yuan, Qiying Yu, Xiaochen Zuo, Ruofei Zhu, Wenyuan Xu, Jiaze Chen, Cheng{-}Xiang Wang, Tiantian Fan, Zhengyin Du, Xiangpeng Wei, Xiangyu Yu, Gaohong Liu, Juncai Liu, Lingjun Liu, Haibin Lin, Zhiqi Lin, Bole Ma, Chi Zhang, Mofan Zhang, Wang Zhang, Hang Zhu, Ru~Zhang, Xin Liu, Mingxuan Wang, Yonghui Wu, and Lin Yan.
\newblock {VAPO:} efficient and reliable reinforcement learning for advanced reasoning tasks.
\newblock \emph{CoRR}, abs/2504.05118, 2025.

\bibitem[Zhang et~al.(2025)Zhang, Yue, Li, Yun, Wan, Wang, Cheng, Yu, and Chen]{Agent-commute}
Guibin Zhang, Yanwei Yue, Zhixun Li, Sukwon Yun, Guancheng Wan, Kun Wang, Dawei Cheng, Jeffrey~Xu Yu, and Tianlong Chen.
\newblock Cut the crap: An economical communication pipeline for llm-based multi-agent systems.
\newblock In \emph{{ICLR}}. OpenReview.net, 2025.

\bibitem[Zhao et~al.(2023{\natexlab{a}})Zhao, Zhou, Li, Tang, Wang, Hou, Min, Zhang, Zhang, Dong, Du, Yang, Chen, Chen, Jiang, Ren, Li, Tang, Liu, Liu, Nie, and Wen]{LLM-survey}
Wayne~Xin Zhao, Kun Zhou, Junyi Li, Tianyi Tang, Xiaolei Wang, Yupeng Hou, Yingqian Min, Beichen Zhang, Junjie Zhang, Zican Dong, Yifan Du, Chen Yang, Yushuo Chen, Zhipeng Chen, Jinhao Jiang, Ruiyang Ren, Yifan Li, Xinyu Tang, Zikang Liu, Peiyu Liu, Jian{-}Yun Nie, and Ji{-}Rong Wen.
\newblock A survey of large language models.
\newblock \emph{CoRR}, abs/2303.18223, 2023{\natexlab{a}}.

\bibitem[Zhao et~al.(2023{\natexlab{b}})Zhao, Gu, Varma, Luo, Huang, Xu, Wright, Shojanazeri, Ott, Shleifer, Desmaison, Balioglu, Damania, Nguyen, Chauhan, Hao, Mathews, and Li]{FSDP}
Yanli Zhao, Andrew Gu, Rohan Varma, Liang Luo, Chien{-}Chin Huang, Min Xu, Less Wright, Hamid Shojanazeri, Myle Ott, Sam Shleifer, Alban Desmaison, Can Balioglu, Pritam Damania, Bernard Nguyen, Geeta Chauhan, Yuchen Hao, Ajit Mathews, and Shen Li.
\newblock Pytorch {FSDP:} experiences on scaling fully sharded data parallel.
\newblock \emph{Proc. {VLDB} Endow.}, 16\penalty0 (12):\penalty0 3848--3860, 2023{\natexlab{b}}.

\bibitem[Zheng et~al.(2025)Zheng, Zhou, Bartoldson, Kailkhura, Lai, Zhao, and Chen]{GRESO}
Haizhong Zheng, Yang Zhou, Brian~R. Bartoldson, Bhavya Kailkhura, Fan Lai, Jiawei Zhao, and Beidi Chen.
\newblock Act only when it pays: Efficient reinforcement learning for {LLM} reasoning via selective rollouts.
\newblock \emph{CoRR}, abs/2506.02177, 2025.

\end{thebibliography}

\newpage
\appendix

\begin{algorithm*}[t]
\centering
\caption{$\operatorname{Pagerank-weighted \ Sequential \ Data \ Pruning}(S, \mathbf{w}, n. k)$}
\label{alg:dpp-sampling}
\begin{minipage}{1\linewidth} 
\small
\begin{algorithmic}
    \Inputs{
        Similarity matrix $S \in \mathbb{R}^{n \times n}$,  
        Pagerank-weighted vector $\mathbf{w} \in \mathbb{R}^n$, 
        Total node $n$, 
        Sample size $k \leq n$ \\
    }
    \Initialize{
        Kernel Matrix $L = \operatorname{diag}(\mathbf{w}^{1/2}) \cdot S \cdot \operatorname{diag}(\mathbf{w}^{1/2})$,
        ${Y} \gets \emptyset$,
        $\mathcal{C} \gets \{1, 2, \dots, n\}$,
        $L = Q\Lambda Q^\top$,
        $\Lambda \gets \operatorname{diag}(\lambda_1, \dots, \lambda_n)$, 
        $V \gets Q \cdot \operatorname{diag}(\lambda_1^{-1/2}, \dots, \lambda_n^{-1/2})$
    }
    \For {$t = 1$ \textbf{to} $k$}
        \State $\mathbf{p} \gets \mathbf{0}_{|\mathcal{C}|}$ \Comment{Initialize zero vector: $\mathbf{p} \in \mathbb{R}^{|\mathcal{C}|}, p_i = 0\ \forall i$}
        \For {$i = 1$ \textbf{to} $n$}
            \If {$i \in \mathcal{C}$}
                \State $\mathbf{p}[i] \gets \left\| V_i \right\|_2^2$
            \EndIf
        \EndFor
        \State $\mathbf{p} \gets \mathbf{p} / \sum_j \mathbf{p}[j]$
        \State $i_t \gets \operatorname{Sample}(\mathcal{C}, \mathbf{p})$ \Comment{$i_t \sim \operatorname{Categorical}(\mathbf{p}), \sum\limits_{j \in \mathcal{C}} p_j=1$}
        \State ${Y} \gets {Y} \cup \{i_t\}$
        \State $\mathcal{C} \gets \mathcal{C} \setminus \{i_t\}$
        \If {$t < k$}
            \State $\mathbf{u} \gets V_{i_t}^\top$ \Comment{Select pivot vector}
            \State $\mathbf{c} \gets \mathbf{u}^{\top} V_{\mathcal{C}}$ 
            \State $V_{\mathcal{C}} \gets V_{\mathcal{C}} - \mathbf{u} \mathbf{c}$  \Comment{Gram-Schmidt orthogonalization}
        \EndIf
    \EndFor
    \State \Return{${Y}$}
\end{algorithmic}
\end{minipage}
\end{algorithm*}

\begin{algorithm*}[t]
\centering
\caption{$\operatorname{Explorability-guided \ Rollout \ Pruning}(\mathcal{B},w,\lambda,\rho,d,e,G,\{O^t\}_{t=1}^e)$}
\label{alg:explorability-pruning}
\begin{minipage}{1\linewidth}
\small
\begin{algorithmic}
\Inputs{
    Raw batch $\mathcal{B} = {(q_i, a_i)}_{i=1}^{|\mathcal{B}|}$,
    Window size $w$,
    Threshold for filtering poor negative rollouts $\lambda$,
    Replay ratio $\rho$,
    Decay rate $d$,
    Current epoch number $e$,
    Rollout numbers per sample $G$,
    Rollout history $\{O^t\}_{t=1}^e$
}
\Initialize{
    Pruned batch $\mathcal{B}^{\text{Pruned}} \gets \emptyset$
}
\For{each sample $(q_i, a_i)$ in $\mathcal{B}$}
    \State $\mathcal{E}_i \gets \mathcal{E}(q_i, a_i, \{O^t_i\}_{t=e-w+1}^e)$ \Comment{Using Equation~\ref{eq:explorability}}
\EndFor
\State $\alpha_e = \alpha_0 - d \cdot e$ \Comment{Calculate high explorability ratios for this epoch}
\State Sort $\mathcal{B}$ in descending order by $\mathcal{E}_i$
\State $\mathcal{B}^{\text{High-Exp}} \gets$ top $\lceil \alpha_e \times |\mathcal{B}| \rceil$ samples from sorted $\mathcal{B}$
\State $\mathcal{B}^{\text{Replay}} \gets$ sample $\lceil \rho \times |\mathcal{B}| \rceil$ samples from $\mathcal{B}$ with the smallest $|\{O^t\}_{t=1}^e|$
\State $\mathcal{B}^{\text{Pruned}} \gets \mathcal{B}^{\text{High}} \cup \mathcal{B}^{\text{Replay}}$
\State \Return $\mathcal{B}^{\text{Pruned}}$

\end{algorithmic}
\end{minipage}
\end{algorithm*}

\section{Theoretical Analysis}
\label{sec:theoretic_proof}

In this part, we provide a theoretical derivation of the optimization objective aimed at maximizing the determinant of a kernel matrix that incorporates both diversity and influence in Equation~\ref{eq:influence_diversity}.

We begin with the original weighted determinant maximization problem:
\begin{align}
\max_{Y \subseteq \mathbf{P}} \left( \det(S_Y) \cdot \prod_{i \in Y} w_i \right),
\end{align}
where $S_Y \in \mathbb{R}^{|Y|\times|Y|}$ denotes the similarity matrix over the subset $Y$, and $w_i$ represents the influential weight (\ie the PageRank score) of each sample $i$.
This objective aims to select a subset $Y$ that is both diverse (as captured by $\det(S_Y)$) and influential (as promoted by the product of weights $\prod_{i \in Y} w_i$).
To combine these two factors more naturally, we express the product of weights in matrix form.
Note that:
\begin{align}
\prod\limits_{i \in Y}w_i = \det (\operatorname{diag} (w_Y)),
\end{align}
where $\operatorname{diag}(w_Y)$ is the diagonal matrix formed by the weights ${w_i}$.
To incorporate the weights directly into the kernel matrix, we consider the square root of the weights.
\begin{align}
\operatorname{diag} (w_Y^{1/2}) = \operatorname{diag}(\sqrt{w_i})_{i \in Y}.
\end{align}
We then observe that:
\begin{align}
\max_{Y \subseteq \mathbf{P}} \left( \det(S_Y) \cdot \prod_{i \in Y} w_i \right) 
&= \max_{Y \subseteq \mathbf{P}} \det\left( \operatorname{diag}(\mathbf{w}_Y^{1/2}) \cdot S_Y \cdot \operatorname{diag}(\mathbf{w}_Y^{1/2}) \right).
\end{align}
This equality follows from the multiplicative property of the determinant and the fact that $\operatorname{diag}(\mathbf{w}_Y^{1/2})$ is a diagonal matrix.
Now, defining the weighted kernel matrix for the subset $Y$ as:
\begin{align}
L_Y = \operatorname{diag}(\mathbf{w}_Y^{1/2}) \cdot S_Y \cdot \operatorname{diag}(\mathbf{w}_Y^{1/2}),
\end{align}
we can rewrite the optimization problem as:
\begin{align}
\max_{Y \subseteq \mathbf{P}} \det(L_Y)
\end{align}
This form corresponds to a standard determinantal point process (DPP) with kernel $L$, where the joint effects of diversity and influence are captured by $L_Y$.

\section{Detailed Experimental Setup}
\label{app:detailed_exp_setup}

\paratitle{Models.}
We run our experiments on DeepSeek-R1-Distill-Qwen-7B~\citep{Deepseek-r1}, DeepSeek-R1-Distill-Llama-8B~\citep{Deepseek-r1}, and Qwen2.5-Math-7B~\citep{Qwen2.5-Math-7B}.
For DeepSeek-R1-Distill-Qwen-7B and Deepseek-R1-Distill-Llama-8B, we set the context length to 16384.
For Qwen2.5-Math-7B models, we set the context length to 4096, as it is the maximum context length for this model.

\paratitle{Training.}
Our method is implemented based on the Verl~\citep{verl} pipeline and uses vLLM~\citep{vllm} for rollout.
We train DeepSeek-R1-Distill-Qwen-7B and DeepSeek-R1-Distill-Llama-8B on 64×H200 GPUs, and Qwen2.5-Math-7B on 32×H200 GPUs.
For training datasets, we use the DAPO-Math~\cite{DAPO} as the training dataset.
During rollouts, we set the temperature to 1 and sample 8 responses per prompt.
The training batch size is set to 256.
We apply the GRPO algorithm to train the model.
Similar to ~\citet{VAPO}, we remove both the KL divergence loss and the entropy loss.
We train all models for 1000 steps, and we optimize the actor model using the AdamW~\citep{AdamW} optimizer with a constant learning rate of 2e-6 for DeepSeek-R1-Distill-Qwen-7B and Deepseek-R1-Distill-Llama-8B and 1e-6 for Qwen-Math-7B.
The actor module is optimized using Fully Sharded Data Parallel (FSDP)~\citep{FSDP} for efficient distributed training.
The chat template we use is ``User: \textbackslash n [question] \textbackslash n Please reason step by step, and put your final answer within \textbackslash boxed\{\}. \textbackslash n \textbackslash n Assistant:''.
For offline data selection, we first apply PageRank-weighted Determinantal Point Process~(DPP) to reduce the sample set to 50\% of its original size.
We then perform difficulty-aware sampling based on a normal distribution to select the final subset, which constitutes 20\% of the full dataset.
The mean $\mu$ and standard deviation $\sigma$ of the difficulty distribution for the final selected subset are set to 0.5 and 0.2, respectively.
In the online data selection phase, we configure the window size of recent epoch $w$ as 5, the initial sampling rate $\alpha_0$ as 1, the decay rate $d$ as 0.05, the replay sample ratio $\rho$ as 0.05, and the threshold $\lambda$ for filtering poor negative rollouts as 1.5.

\paratitle{Evaluation.}
For evaluation benchmarks, we use three widely used complex mathematical reasoning benchmarks (\ie AIME24, AIME25, Math500~\citep{Math500}) and two other reasoning benchmarks (\ie GPQA~\citep{GPQA} and LiveCodeBench~\citep{LiveCodeBench}) to evaluate the model performance.
We follow ~\citet{GRESO} to evaluate models on those benchmarks every 50 steps and report the performance of the checkpoint that obtains the best average performance on five benchmarks.
All evaluations are conducted in a zero-shot setting.
We evaluate all models using temperature = 1 and repeat the test set 32 times for evaluation stability for all benchmarks and report the average accuracy.
\section{Detailed Description of Baselines}
\label{app:baselines}

In this part, we provide detailed descriptions of all the baselines used in our experiments.
For offline data selection methods, we compare our method with random selection, conventional supervised fine-tuning~(SFT) data selection methods (\ie PPL-Top~\citep{PPL-Top} and PPL-Middle~\citep{PPL-Middle}), RLVR selection method (\ie LIMR~\citep{LIMR} and Learnalign~\citep{Learnalign}).

$\bullet$ \textbf{\underline{Random}}:
Randomly samples data from the training set.

$\bullet$ \textbf{\underline{PPL-Top}}~\citep{PPL-Top}:
Selects the data with the highest perplexity.

$\bullet$ \textbf{\underline{PPL-Middle}}~\citep{PPL-Middle}:
selects the data with the middle perplexity.

$\bullet$ \textbf{\underline{LIMR}}~\citep{LIMR}:
selects the data whose learning patterns complement the model’s overall
reward trajectory.

$\bullet$ \textbf{\underline{Learnalign}}~\citep{Learnalign}:
selects the data based on representativeness (measured via gradients during warmup training) and difficulty (determined by rollout accuracy).

For online data selection methods, we incorporate them into our offline selected subset and compare against random online selection and GRESO~\citep{GRESO}.

$\bullet$ \textbf{\underline{Random}}:
Randomly filters 40\% of the data at each batch prior to rollout during training.

$\bullet$ \textbf{\underline{GRESO}}~\citep{GRESO}:
Probabilistically filter historical samples with zero variance at each batch before rollout during training.
\section{Additional Experiments}


\subsection{Additional Hyperparameter Analysis}

\begin{wrapfigure}{r}{0.5\textwidth}
    \centering
    \begin{subfigure}[b]{0.48\linewidth}
        \centering
        \includegraphics[width=\linewidth]{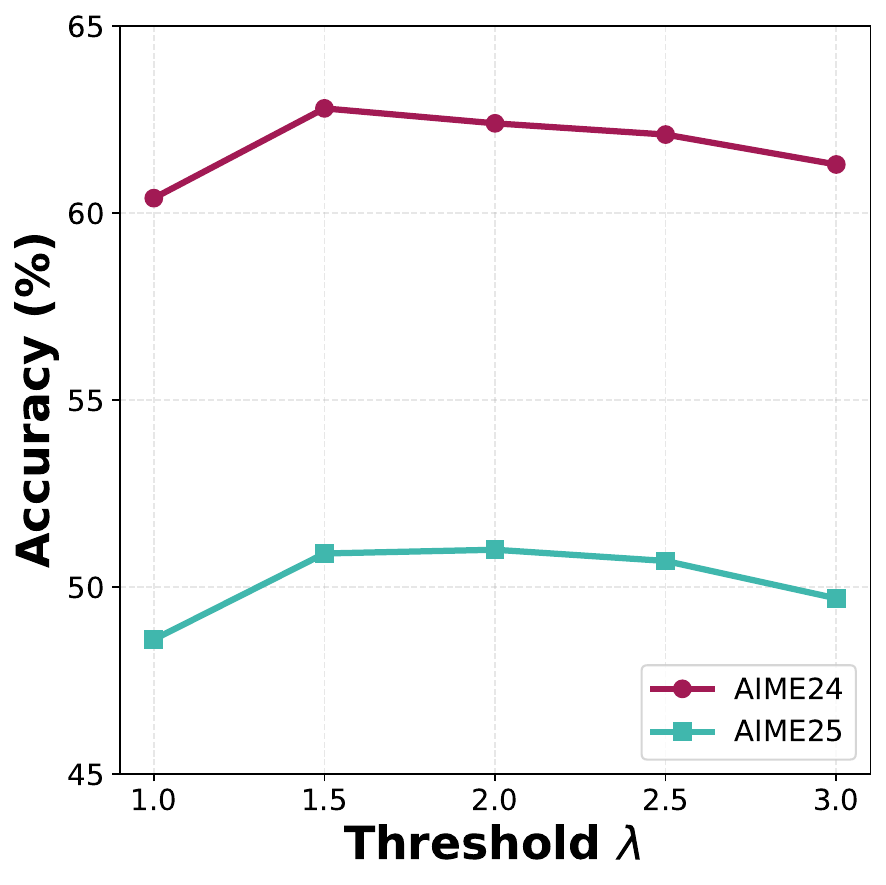}
        \caption{Threshold}
        \label{fig:np_ratio}
    \end{subfigure}
    \hfill
    \begin{subfigure}[b]{0.48\linewidth}
        \centering
        \includegraphics[width=\linewidth]{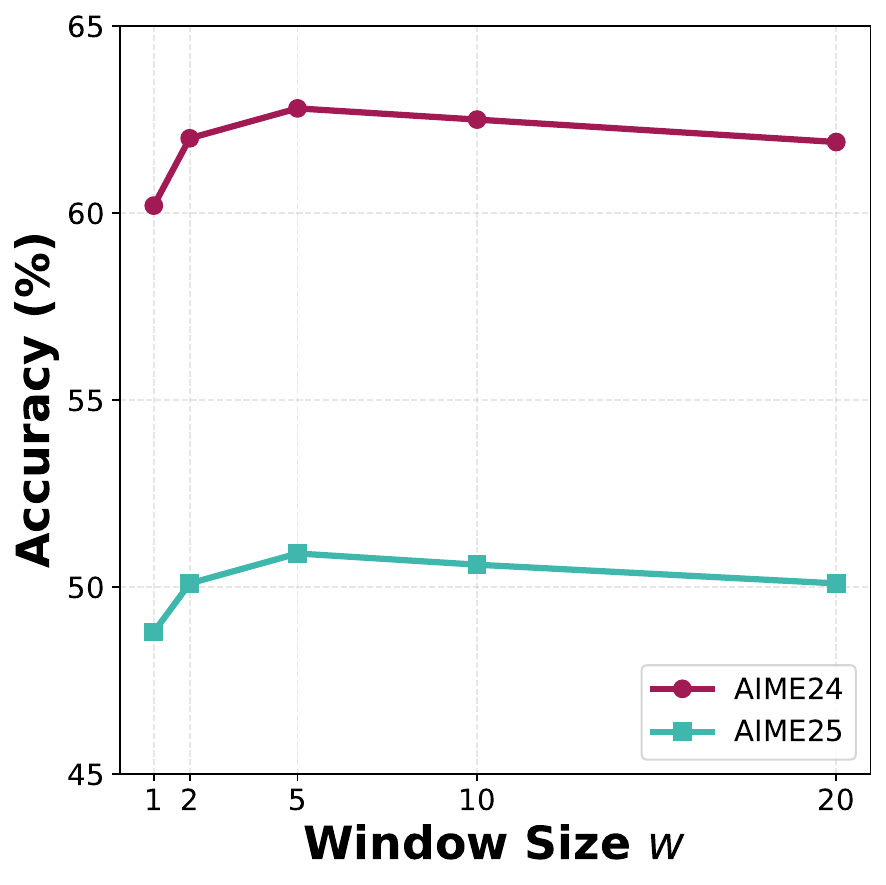}
        \caption{Window size}
        \label{fig:window_size}
    \end{subfigure}
    \caption{Hyper Parameter Analysis}
    \label{fig:hyper_para2}
\end{wrapfigure}

In this section, we present further hyperparameter analysis on the threshold $\lambda$ for high-quality negative rollouts and the window size $w$.
As shown in Figure~\ref{fig:np_ratio}, performance first improves and then declines as the threshold increases.
This indicates that when the threshold is set too low, potentially useful samples that could help exploration might be excluded.
Conversely, an excessively high threshold may lead to the inclusion of unreasonable or noisy samples (\eg nonsensical text rollouts), which can adversely affect model performance.
Regarding the window size $w$, Figure~\ref{fig:window_size} indicates that the model performs best when $w = 5$.
This suggests that the window size should be chosen within an appropriate range.
One possible explanation is that a very small window may not capture broader historical training dynamics, while an overly large window may not focus on recent training trends.
\section{Case Study}

In this section, we provide detailed case studies on five reasoning benchmarks.

\lstset{
    breaklines=true,
    breakindent=0pt,
    breakautoindent=false,
}

\begin{center}
\begin{tcolorbox}[colback=blue!5!white,colframe=blue!55!black,width=\linewidth,title={AIME24}, fontupper=\small]
    {
    \large
    \textbf{[Question:]}

    \small
    \begin{lstlisting}
Every morning Aya goes for a $9$-kilometer-long walk and stops at a coffee shop afterwards. When she walks at a constant speed of $s$ kilometers per hour, the walk takes her 4 hours, including $t$ minutes spent in the coffee shop. When she walks $s+2$ kilometers per hour, the walk takes her 2 hours and 24 minutes, including $t$ minutes spent in the coffee shop. Suppose Aya walks at $s+\\frac{1}{2}$ kilometers per hour. Find the number of minutes the walk takes her, including the $t$ minutes spent in the coffee shop.\nPlease reason step by step, and put your final answer within \\boxed{{}}.
    \end{lstlisting}

    \large
    \textbf{[Answer:]}

    \small
    \begin{lstlisting}
Okay, so I have this problem about Aya's morning walk and her coffee stops. Let me try to figure this out step by step.\n\nFirst, let's parse the information given:\n\n1. Every morning, Aya goes for a 9-kilometer-long walk and then stops at a coffee shop.\n2. When she walks at a constant speed of 's' kilometers per hour, the entire walk, including her coffee stop, takes her 4 hours. Wait, hold on, the problem says it takes her 4 hours, but 4 hours is 240 minutes. But then it mentions including t minutes spent in the coffee shop. Hmm, maybe I need to clarify.\n\nWait, maybe the total time includes both walking and coffee time. So, when she walks at speed s km/h, her total time is 4 hours, which includes t minutes in the coffee shop.\n\nSimilarly, when she walks at s + 2 km/h, the total time is 2 hours and 24 minutes, which also includes t minutes coffee time.\n\nOur goal is to find the time it takes her when she walks at s + 1/2 km/h, again including the t minutes coffee stop.

...

**Final Answer**\n\\boxed{204}\n</think>\n\nGiven that Aya goes for a 9-kilometer-long walk every morning and stops at a coffee shop, we need to determine the total time she takes when she walks at a speed of \\( s + \\frac{1}{2} \\) km/h, including the time spent in the coffee shop.\n\nFirst, we set up the equations based on the given information:\n\n1. When walking at speed \\( s \\) km/h:\n   - Total time = 4 hours = 240 minutes.\n   - Let \\( t \\) be the time spent in the coffee shop.\n   - Equation: \\( \\frac{540}{s} + t = 240 \\).\n\n2. When walking at speed \\( s + 2 \\) km/h:\n   - Total time = 2 hours 24 minutes = 144 minutes.\n   - Equation: \\( \\frac{540}{s + 2} + t = 144 \\).\n\nSubtracting the second equation from the first to eliminate \\( t \\):\n\\[ \\frac{540}{s} - \\frac{540}{s + 2} = 96 \\]\n\\[ 540 \\left( \\frac{1}{s} - \\frac{1}{s + 2} \\right) = 96 \\]\n\\[ 540 \\left( \\frac{2}{s(s + 2)} \\right) = 96 \\]\n\\[ \\frac{1080}{s(s + 2)} = 96 \\]\n\\[ \\frac{1080}{96} = s(s + 2) \\]\n\\[ 11.25 = s^2 + 2s \\]\n\\[ s^2 + 2s - 11.25 = 0 \\]\n\nSolving the quadratic equation:\n- Discriminant: \\( 4 + 45 = 49 \\)\n- Solutions: \\( s = \\frac{-2 \\pm 7}{2} \\)\n- Positive solution: \\( s = 2.5 \\) km/h.\n\nUsing \\( s = 2.5 \\) to find \\( t \\):\n- From \\( \\frac{540}{2.5} + t = 240 \\):\n  - \\( 216 + t = 240 \\)\n  - \\( t = 24 \\) minutes.\n\nWhen walking at \\( s + \\frac{1}{2} = 3 \\) km/h:\n- Time to walk 9 km: \\( \\frac{9}{3} = 3 \\) hours = 180 minutes.\n- Total time including coffee: \\( 180 + 24 = 204 \\) minutes.\n\nThus, the total time taken is \\(\\boxed{204}\\) minutes.
    \end{lstlisting}
    
    }
    
\end{tcolorbox}
\end{center}
\lstset{
    breaklines=true,
    breakindent=0pt,
    breakautoindent=false,
}

\begin{center}
\begin{tcolorbox}[colback=blue!5!white,colframe=blue!55!black,width=\linewidth,title={AIME25}]
    {
    \large
    \textbf{[Question:]}

    \small
    \begin{lstlisting}
Find the sum of all integer bases $b>9$ for which $17_b$ is a divisor of $97_b.$\nPlease reason step by step, and put your final answer within \\boxed{{}}.
    \end{lstlisting}

    \large
    \textbf{[Answer:]}

    \small
    \begin{lstlisting}
Alright, let's see. I have this problem here: I need to find the sum of all integer bases \\( b > 9 \\) for which \\( 17_b \\) is a divisor of \\( 97_b \\). Hmm, okay. So, bases greater than 9, which means we're looking at bases starting from 10 upwards, right? \n\nFirst, I remember that when numbers are written in a certain base \\( b \\), their value in decimal (base 10) can be calculated. For example, a number like \\( 17_b \\) is equal to \\( 1 \\times b + 7 \\times 1 = b + 7 \\) in base 10. Similarly, \\( 97_b \\) would be \\( 9 \\times b + 7 \\times 1 = 9b + 7 \\) in base 10. \n\nSo, the problem is asking for all bases \\( b > 9 \\) where \\( 17_b \\) divides \\( 97_b \\). That means that when we convert both numbers to base 10, \\( 97_b \\) should be divisible by \\( 17_b \\) in base 10 terms.

...

**Final Answer**\n\\boxed{70}\n</think>\n\nTo solve the problem, we need to find all integer bases \\( b > 9 \\) for which \\( 17_b \\) is a divisor of \\( 97_b \\). \n\nFirst, we convert the numbers from base \\( b \\) to base 10:\n- \\( 17_b = 1 \\times b + 7 = b + 7 \\)\n- \\( 97_b = 9 \\times b + 7 = 9b + 7 \\)\n\nWe need \\( b + 7 \\) to divide \\( 9b + 7 \\). This translates to the congruence:\n\\[ 9b + 7 \\equiv 0 \\pmod{b + 7} \\]\n\nRewriting this, we let \\( n = b + 7 \\), so \\( b = n - 7 \\). Substituting back, we get:\n\\[ 9(n - 7) + 7 \\equiv 0 \\pmod{n} \\]\n\\[ 9n - 63 + 7 \\equiv 0 \\pmod{n} \\]\n\\[ 9n - 56 \\equiv 0 \\pmod{n} \\]\n\\[ -56 \\equiv 0 \\pmod{n} \\]\nThis implies \\( n \\) is a divisor of 56. Since \\( b > 9 \\), \\( n = b + 7 > 16 \\). The divisors of 56 greater than 16 are 28 and 56.\n\nThus, solving for \\( b \\):\n- If \\( n = 28 \\), then \\( b = 28 - 7 = 21 \\)\n- If \\( n = 56 \\), then \\( b = 56 - 7 = 49 \\)\n\nBoth 21 and 49 are valid bases as they satisfy the digit constraints in their respective numbers. Therefore, the sum of these bases is:\n\\[ 21 + 49 = 70 \\]\n\n\\[\n\\boxed{70}\n\\]
    \end{lstlisting}
    
    }
    
\end{tcolorbox}
\end{center}
\lstset{
    breaklines=true,
    breakindent=0pt,
    breakautoindent=false,
}

\begin{center}
\begin{tcolorbox}[colback=blue!5!white,colframe=blue!55!black,width=\linewidth,title={MATH500}]
    {
    \large
    \textbf{[Question:]}

    \small
    \begin{lstlisting}
Define\n\\[p = \\sum_{k = 1}^\\infty \\frac{1}{k^2} \\quad \\text{and} \\quad q = \\sum_{k = 1}^\\infty \\frac{1}{k^3}.\\]Find a way to write\n\\[\\sum_{j = 1}^\\infty \\sum_{k = 1}^\\infty \\frac{1}{(j + k)^3}\\]in terms of $p$ and $q.$\nPlease reason step by step, and put your final answer within \\boxed{}.
    \end{lstlisting}

    \large
    \textbf{[Answer:]}

    \small
    \begin{lstlisting}
Okay, so I have this problem here where I need to express the double sum \\(\\sum_{j = 1}^\\infty \\sum_{k = 1}^\\infty \\frac{1}{(j + k)^3}\\) in terms of \\(p\\) and \\(q\\), where \\(p = \\sum_{k = 1}^\\infty \\frac{1}{k^2}\\) and \\(q = \\sum_{k = 1}^\\infty \\frac{1}{k^3}\\). Hmm, interesting. I remember that \\(p\\) is related to the Basel problem and equals \\(\\frac{\\pi^2}{6}\\), and \\(q\\) is Apery's constant, approximately 1.202... But maybe I don't need their exact values, just to express the double sum in terms of \\(p\\) and \\(q\\).\n\nSo, the double sum is over two indices, \\(j\\) and \\(k\\), both starting from 1 to infinity, of \\(1/(j + k)^3\\). I think I need to find a clever way to rewrite this or maybe relate it to the known series \\(p\\) and \\(q\\). Let's think about how to handle double sums. Often, double sums can sometimes be converted into single sums by clever manipulation or using integral representations.\n\nLet me write down the double sum explicitly:\n\n\\[\nS = \\sum_{j=1}^\\infty \\sum_{k=1}^\\infty \\frac{1}{(j + k)^3}\n\\]\n\nI recall that for such sums, sometimes we can change variables to simplify the expression. Maybe set \\(n = j + k\\).

...

For each \\(n\\), \\(j\\) can range from 1 to \\(n - 1\\), so we rewrite the double sum as a single sum:\n\n\\[\nS = \\sum_{j=1}^\\infty \\sum_{k=1}^\\infty \\frac{1}{(j + k)^3} = \\sum_{n=2}^\\infty \\sum_{j=1}^{n - 1} \\frac{1}{n^3}\n\\]\n\nFor each \\(n\\), there are \\(n - 1\\) pairs \\((j, k)\\), so:\n\n\\[\nS = \\sum_{n=2}^\\infty (n - 1) \\cdot \\frac{1}{n^3}\n\\]\n\nWe split the fraction:\n\n\\[\n\\frac{n - 1}{n^3} = \\frac{1}{n^2} - \\frac{1}{n^3}\n\\]\n\nThus,\n\n\\[\nS = \\sum_{n=2}^\\infty \\left( \\frac{1}{n^2} - \\frac{1}{n^3} \\right)\n\\]\n\nThis can be written as:\n\n\\[\nS = \\left( \\sum_{n=2}^\\infty \\frac{1}{n^2} \\right) - \\left( \\sum_{n=2}^\\infty \\frac{1}{n^3} \\right)\n\\]\n\nUsing the known series \\(p = \\sum_{n=1}^\\infty \\frac{1}{n^2}\\) and \\(q = \\sum_{n=1}^\\infty \\frac{1}{n^3}\\), we have:\n\n\\[\n\\sum_{n=2}^\\infty \\frac{1}{n^2} = p - 1 \\quad \\text{and} \\quad \\sum_{n=2}^\\infty \\frac{1}{n^3} = q - 1\n\\]\n\nTherefore,\n\n\\[\nS = (p - 1) - (q - 1) = p - q\n\\]\n\nThe final answer is: \\boxed{p - q} 
    \end{lstlisting}
    
    }
    
\end{tcolorbox}
\end{center}
\lstset{
    breaklines=true,
    breakindent=0pt,
    breakautoindent=false,
}

\begin{center}
\begin{tcolorbox}[colback=blue!5!white,colframe=blue!55!black,width=\linewidth,title={GPQA}]
    {
    \large
    \textbf{[Question:]}

    \small
    \begin{lstlisting}
Answer the following multiple choice question. The last line of your response should be of the following format: 'Answer: $LETTER' (without quotes) where LETTER is one of ABCD. Think step by step before answering.\n\nTwo quantum states with energies E1 and E2 have a lifetime of 10^-9 sec and 10^-8 sec, respectively. We want to clearly distinguish these two energy levels. Which one of the following options could be their energy difference so that they can be clearly resolved?\n\n\nA) 10^-9 eV\nB) 10^-11 eV\nC) 10^-8 eV\n\nD) 10^-4 eV
    \end{lstlisting}

    \large
    \textbf{[Answer:]}

    \small
    \begin{lstlisting}

Okay, so I have this quantum mechanics problem in front of me. Hmm, let's try to think it through. The question is about distinguishing two quantum states with different lifetimes. Right, I remember that in quantum mechanics, the uncertainty principle plays a role in determining the observable differences between energy levels. Or wait, perhaps it's about the Heisenberg uncertainty principle? Or maybe it's related to the line width due to the energy uncertainty affecting the transition between states.\n\nSo the problem states that two quantum states have energies E1 and E2 with lifetimes of 1e-9 seconds and 1e-8 seconds, respectively. The goal is to find which energy difference would allow us to clearly distinguish these two energy levels. The options are A to D.\n\nI'm a bit rusty on this, but I think the key here is the concept of the spectral line width. Oh wait, I remember something about the width of spectral lines being related to the uncertainty in energy, or perhaps the Heisenberg uncertainty principle. Because if the lifetimes are different, their energy differences might lead to different spectral lines that can be resolved.\n\nHeisenberg's uncertainty principle says that the product of the uncertainty in energy (\delta E) and the uncertainty in time (\deltat) is approximately h/2.

...

For the two states to be clearly resolved, the energy difference \\( \\Delta E \\) between them must be greater than the sum of their linewidths:\n\n\\[\n\\Delta E > \\Gamma_1 + \\Gamma_2 \\approx 1.316 \\times 10^{-7} \\, \\text{eV}.\n\\]\n\nNow, we evaluate the given options:\n\n- **Option A: \\( 10^{-9} \\, \\text{eV} \\)**  \n  This is much smaller than \\( 1.316 \\times 10^{-7} \\, \\text{eV} \\), so it does not satisfy the condition.\n\n- **Option B: \\( 10^{-11} \\, \\text{eV} \\)**  \n  This is even smaller than \\( 10^{-9} \\, \\text{eV} \\), so it does not satisfy the condition.\n\n- **Option C: \\( 10^{-8} \\, \\text{eV} \\)**  \n  This is \\( 1.0 \\times 10^{-8} \\, \\text{eV} \\), which is smaller than \\( 1.316 \\times 10^{-7} \\, \\text{eV} \\), so it does not satisfy the condition.\n\n- **Option D: \\( 10^{-4} \\, \\text{eV} \\)**  \n  This is \\( 1.0 \\times 10^{-4} \\, \\text{eV} \\), which is much larger than \\( 1.316 \\times 10^{-7} \\, \\text{eV} \\), so it satisfies the condition.\n\nThus, the only energy difference that allows the two states to be clearly resolved is \\( 10^{-4} \\, \\text{eV} \\).\n\nAnswer: D
    \end{lstlisting}
    
    }
    
\end{tcolorbox}
\end{center}
\lstset{
    breaklines=true,
    breakindent=0pt,
    breakautoindent=false,
}

\begin{center}
\begin{tcolorbox}[colback=blue!5!white,colframe=blue!55!black,width=\linewidth,title={LiveCodeBench}]
    {
    \large
    \textbf{[Question:]}

    \small
    \begin{lstlisting}
### Question:\nYou are given an integer sequence A=(A_1,\\ldots,A_N) of length N. Here, A_1, A_2, \\ldots, A_N are all distinct.\nWhich element in A is the second largest?\n\nInput\n\nThe input is given from Standard Input in the following format:\nN \r\nA_1 A_2 \\ldots A_{N}\n\nOutput\n\nPrint the integer X such that the X-th element in A is the second largest.\n\nConstraints\n\n\n- 2 \\leq N \\leq 100\n- 1 \\leq A_i \\leq 10^9\n- A_1, A_2, \\ldots, A_N are all distinct.\n- All input values are integers.\n\nSample Input 1\n\n4\r\n8 2 5 1\n\nSample Output 1\n\n3\r\n\nThe second largest element in A is A_3, so print 3.\n\nSample Input 2\n\n8\r\n1 2 3 4 5 10 9 11\n\nSample Output 2\n\n6\n\n### Format: Read the inputs from stdin solve the problem and write the answer to stdout (do not directly test on the sample inputs). Enclose your code within delimiters as follows.\n```python\n# YOUR CODE HERE\n```\n\n### Answer: (use the provided format with backticks)\n\n
    \end{lstlisting}

    \large
    \textbf{[Answer:]}

    \small
    \begin{lstlisting}

Okay, I need to solve this problem where I have to find the second largest element in a given integer sequence and then determine which position it occupies in the list, specifically the index in the 1-based notation. Hmm, let's think through how to approach this.\n\nWait, wait. The problem says that each element is distinct, so I don't have to worry about duplicates. The task is to find the second largest number and then output the index (but wait, the problem says the X-th element, which is the position in the array, not zero-based). Oh right, the first sample input: 8 2 5 1. The largest is 8, then 5, then 2, then 1. Wait no, wait the second sample input is 1 2 3 4 5 10 9 11. The largest is 11, then 10, and then 9. Wait, no, wait in sample input 2, the array is [1,2,3,4,5,10,9,11], the largest is 11 at position 8, the second largest is 10 at position 6. So the output is 6. So the task is to find the second largest element in the list, which is the element that is just smaller than the maximum, but it could be more than just that because there might be elements that are after the maximum which are smaller but larger than the rest.

...

### Solution Code\n```python\nn = int(input())\na = list(map(int, input().split()))\nmax_val = max(a)\ncandidates = [x for x in a if x != max_val]\nsecond_val = max(candidates)\nindex = a.index(second_val) + 1  # Convert to 1-based index\nprint(index)\n```\n\n### Explanation\n- **Reading Input**: We read the integer `n` which is the length of the list and then read the list of integers `a`.\n- **Finding Maximum**: Using Python's built-in `max()` function, we determine the largest value in the list.\n- **Filtering Candidates**: We create a new list `candidates` that excludes the maximum value to focus on the remaining elements.\n- **Finding Second Largest**: The maximum of the `candidates` list is the second largest element in the original list.\n- **Index Conversion**: The index of this second largest element is found using `index()`, which returns a 0-based index, so we add 1 to convert it to 1-based indexing before printing the result.\n\nThis approach ensures that we efficiently find the second largest element and correctly determine its position in the original list, adhering to the problem constraints and requirements.
    \end{lstlisting}
    
    }
    
\end{tcolorbox}
\end{center}

\end{document}